\documentclass{article}

\usepackage{microtype}
\usepackage{graphicx}
\usepackage{subfigure}
\usepackage{booktabs} % for professional tables
\usepackage{nidanfloat}

\usepackage{hyperref}

\usepackage[accepted]{icml2019}

\usepackage[T1]{fontenc}
\usepackage[utf8]{inputenc}

\usepackage{algorithm}
\usepackage{amsmath}
\usepackage{amssymb}
\usepackage{amsthm}
\usepackage{bm}
\usepackage{cleveref}
\usepackage{dsfont}
\usepackage{float}
\usepackage{markdown}
\usepackage{mathabx}
\usepackage{mathtools}
\usepackage{mleftright}
\usepackage{moresize}
\usepackage{multirow}
\usepackage{nicefrac}
\usepackage{thmtools}
\usepackage{wrapfig}
\usepackage{xcolor}
\usepackage{zref}

\newcommand*{\alab}[1]{\label{alg:#1}}
\newcommand*{\aref}[1]{Algorithm~\ref{alg:#1}}
\newcommand*{\elab}[1]{\label{eq:#1}}
\newcommand*{\eref}[1]{\eqref{eq:#1}}
\newcommand*{\flab}[1]{\label{fig:#1}}
\newcommand*{\fref}[1]{Figure~\ref{fig:#1}}
\newcommand*{\slab}[1]{\label{sec:#1}}
\newcommand*{\sref}[1]{Section~\ref{sec:#1}}

\newcommand*{\xlab}[1]{\label{apx:#1}}
\newcommand*{\xref}[1]{Appendix~\ref{apx:#1}}

\makeatletter\@ifpackageloaded{algorithmic}{}{
\usepackage[noend]{algpseudocode}
\renewcommand\algorithmiccomment[1]{\quad// #1}
\algblockdefx{Prob}{EndProb}[1]{\textbf{With probability} #1}{}
\algcblockdefx{Prob}{Else}{EndProb}{\textbf{Otherwise}}{}
}\makeatother

\newcommand*{\B}[1]{\ifmmode\mathbf{#1}\else\textbf{#1}\fi}
\newcommand*{\C}[1]{\mathcal{#1}}
\newcommand*{\F}[1]{\mathbb{#1}}

\newcommand*{\Z}[1]{\mathds{#1}}

\DeclareMathOperator{\D}{\Z D}

\DeclareMathOperator{\dif}{\partial}

\DeclareMathOperator{\E}{\Z E}

\newcommand*{\const}{\textnormal{const}}
\newcommand*{\copr}{\pagestyle{fancy}\thispagestyle{fancy}\renewcommand{\headrulewidth}{0pt}\lhead{}\rhead{}\rfoot{\tiny Roy Fox, \the\year}}

\newcommand*{\eq}[1]{\begin{align*}#1\end{align*}}

\newcommand*{\eqn}[1]{\begin{align}#1\end{align}}

\newcommand*{\nn}{\nonumber}
\newcommand*{\nul}{\emptyset}

\renewcommand*{\vdots}{\vbox{\baselineskip=4pt\lineskiplimit=0pt\kern2pt\hbox{.}\hbox{.}\hbox{.}}}

\makeatletter
\def\fn{\gdef\@thefnmark{}\@footnotetext}
\makeatother

\makeatletter
\zref@newprop{mzheight}[0pt]{\the\ht\z@}
\zref@newprop{mzdepth}[0pt]{\the\dp\z@}
\newcount\c@@mz
\newcommand*{\the@mz}{mz\the\c@@mz}
\newcommand*{\@mz@list}{}    
\let\@mz@do\relax
\newcommand*{\mzreset}{%
  \begingroup
    \def\@mz@do##1{%
      \global\expandafter\let\csname mz@##1\endcsname\relax
    }%
    \@mz@list
    \global\let\@mz@list\@empty
  \endgroup
}
\newcommand*{\mzleft}[3]{%
  \@ifundefined{mz@#1}{%
    \global\advance\c@@mz\@ne
    \expandafter\xdef\csname mz@#1\endcsname{\the@mz}%
    \xdef\@mz@list{\@mz@list\@mz@do{#1}}%
  }{}%
  \expandafter\let\expandafter\@mz\csname mz@#1\endcsname
  \mleft#2%
  \expandafter\mathpalette\expandafter{%
    \expandafter\@mzleft\expandafter{\@mz}%
  }{#3}%
  \mright.\kern-\nulldelimiterspace
}
\newcommand*{\mzright}[3]{%
  \kern-\nulldelimiterspace
  \@ifundefined{mz@#1}{%   
    \@latex@warning{Missing \string\mzleft{#1}}%
    \mleft.#2\mright#3%
  }{%
    \expandafter\let\expandafter\@mz\csname mz@#1\endcsname
    \mleft.%
    \expandafter\mathpalette\expandafter{%
      \expandafter\@mzright\expandafter{\@mz}%
    }{#2}%
    \mright#3%
  }%
}   
\newcommand*{\@mzleft}{%
  \@mzleftright lr%
}
\newcommand*{\@mzright}{%
  \@mzleftright rl%
}
\newcommand*{\@mzleftright}[5]{%
  \sbox0{$\m@th#4{}#5{}$}%
  \ifmeasuring@
  \else
    \begingroup
      \let\@auxout\@mainaux
      \zref@labelbyprops{#3#1}{mzheight,mzdepth}%
    \endgroup
  \fi
  \zifrefundefined{\@mz #2}{%
  }{%
    \dimen@=\zref@extract{#3#2}{mzheight}\relax
    \ifdim\dimen@>\ht0 %
      \ht0=\dimen@
    \fi
    \dimen@=\zref@extract{#3#2}{mzdepth}\relax
    \ifdim\dimen@>\dp0 %
      \dp0=\dimen@
    \fi
  }%   
  \copy0\relax
}
\makeatother

\begin{document}

\twocolumn[
\icmltitle{Hierarchical Variational Imitation Learning of Control Programs}



\icmlsetsymbol{equal}{*}

\begin{icmlauthorlist}
\icmlauthor{Roy Fox}{uci}
\icmlauthor{Richard Shin}{eecs}
\icmlauthor{William Paul}{eecs}
\icmlauthor{Yitian Zou}{eecs}

\icmlauthor{Dawn Song}{eecs}
\icmlauthor{Ken Goldberg}{eecs,ieor}
\icmlauthor{Pieter Abbeel}{eecs}
\icmlauthor{Ion Stoica}{eecs}
\end{icmlauthorlist}

\icmlaffiliation{uci}{Department of Computer Science, University of California, Irvine}
\icmlaffiliation{eecs}{Department of Electrical Engineering and Computer Sciences, University of California, Berkeley}
\icmlaffiliation{ieor}{Department of Industrial Engineering and Operations Research, University of California, Berkeley}

\icmlcorrespondingauthor{Roy Fox}{royf@uci.edu}


\vskip 0.3in
]



\printAffiliationsAndNotice{}  % leave blank if no need to mention equal contribution

\begin{abstract}
Autonomous agents can learn by imitating teacher demonstrations of the intended behavior. Hierarchical control policies are ubiquitously useful for such learning, having the potential to break down structured tasks into simpler sub-tasks, thereby improving data efficiency and generalization. In this paper, we propose a variational inference method for imitation learning of a control policy represented by parametrized hierarchical procedures (PHP), a program-like structure in which procedures can invoke sub-procedures to perform sub-tasks. Our method discovers the hierarchical structure in a dataset of observation--action traces of teacher demonstrations, by learning an approximate posterior distribution over the latent sequence of procedure calls and terminations. Samples from this learned distribution then guide the training of the hierarchical control policy. We identify and demonstrate a novel benefit of variational inference in the context of hierarchical imitation learning: in decomposing the policy into simpler procedures, inference can leverage acausal information that is unused by other methods. Training PHP with variational inference outperforms LSTM baselines in terms of data efficiency and generalization, requiring less than half as much data to achieve a 24\% error rate in executing the bubble sort algorithm, and to achieve no error in executing Karel programs.
\end{abstract}
\section{Introduction}

Autonomous agents that interact with their environment can learn to perform desired tasks when provided with informative learning signals.
In imitation learning (IL), a teacher provides demonstrations of successful performance of the task, which consist of traces of sensory observations and the desired control actions during execution of the control policy of the teacher~\cite{schaal1999imitation, argall2009survey, gail, hussein2017imitation} or of the learner~\cite{dagger, aggrevate, aggrevated}.
This places some burden on the teacher to correctly control the system, requiring at least an implicit knowledge of \emph{how} to perform the task, rather than just \emph{what} the task is.
However, such implicit knowledge is in many cases more available to humans than a formal description of success criteria or a reward function --- some things are ``easier done than said'', e.g. household chores or grammatical speech.
Moreover, the rich supervision signal in IL is generally more informative than the reward signal provided in reinforcement learning (RL), potentially reducing the required amount of interaction with the system and the teacher.

Structured control, and in particular hierarchical control, has the potential to further improve data efficiency.
Modularity and hierarchy facilitate specialization and abstraction, so that each distinct module of the controller can focus on simpler behavior that is useful in a subset of system states.
This specialization reduces the complexity of the relevant features of the state and of the environment dynamics, allowing each module to have a simpler model that can be learned from less data.

This paper takes a hierarchical imitation learning (HIL) approach, by modeling the control policy as parametrized hierarchical procedures (PHP)~\cite{fox2018parametrized}, a program-like structure in which each procedure, in each step it takes, can either invoke a sub-procedure, take a control action, or terminate and return to its caller.
Given a dataset of observation--action traces of the execution of a teacher control policy, we seek to train the parameters of all procedures in a learner control policy.
Taken together, these trained procedures form a control program, with a conditional branching structure induced by each procedure's choice of sub-procedure calls.
This structure is latent in the data and must be discovered, which is particularly challenging with deep hierarchies where multiple levels of nested procedures can call each other.

\iffalse
PHPs can be formulated as a recurrent neural network (RNN) with discrete memory structure.
Each memory update consists of a sequence of procedure calls and terminations, until a control action is selected.
In order to learn PHPs with gradient-based optimization methods, we represent each PHP as a neural network outputting a distribution over the union set of sub-procedures, control actions, and the termination indicator.
Thus parametrized, PHPs are stochastic RNNs.
\fi

We propose hierarchical variational imitation learning (HVIL): imitation learning of hierarchical control policies via variational inference.
We train an inference model to approximate the posterior distribution, given a demonstration, of the latent sequence of procedure calls and terminations that could generate that trace.
We sample from this distribution to impute the latent variables and guide the training of the generative PHP.

We experiment with our method to train control programs in two domains.
In the Bubble Sort domain, we learn a policy that performs bubble sort of a memory array.
Using HVIL to train a PHP with a 4-level hierarchy of 6 procedures, we outperform a 4-layer LSTM baseline in data efficiency, requiring less than half as much data to achieve a 24\% error rate.
In the Karel domain, we learn to imitate the execution of programs in the Karel language.
On some programs, PHP trained with HVIL achieves zero test errors with less than half as much data as an LSTM baseline.
PHP also generalize better than LSTMs to longer executions than seen in training.

Our work extends SRNNs~\cite{fraccaro2016sequential} to hierarchically-structured discrete latent variables, making three contributions.
First, we present a network architecture for an inference model of a hierarchical control policy.
Second, we provide analysis of technical considerations that we found necessary for training discrete stochastic RNNs, namely analytic KL and Rao--Blackwellization of non-reparametrizable latent variables.
Third, we identify and demonstrate a novel benefit of variational inference in the context of hierarchical imitation learning, leveraging acausal information unused by other methods to facilitate the decomposition of the control policy into simpler procedures.
The code used in our experiments is available at \url{https://github.com/royf/hvil}.

\section{Related Work}

Frameworks of hierarchical control often include explicit or implicit call-stacks.
In the popular options framework~\cite{sutton1999between}, options can use ``intra-option'' control to call other options.
StackRNNs~\cite{joulin2015inferring} can explicitly perform stack operations as part of their control.
Neural Programmers--Interpreters (NPI)~\cite{reed2015neural} control program execution by maintaining a call-stack of procedures and their arguments.
Neural Program Lattices~\cite{li2017neural} add the ability to train NPI when the hierarchical structure is latent in some of the traces, using lattices to group the exponentially many latent stack trajectories into a manageable set.
Parametrized hierarchical procedures (PHP)~\cite{fox2018parametrized} maintains a similar call-stack, and trains it with a level-by-level application of an exact-inference method.
While StackRNN and NPI keep real vectors on the stack, call-stacks of options and PHP maintain discrete values, namely the \emph{identifiers} of the called options or procedures.
PHP, being inspired by procedural programming, differs from options by additionally maintaining a program counter for each procedure.

In this work, we propose to train PHP via a variational inference (VI) method.
VI has been used for training autoencoders in unsupervised learning~\cite{kingma2013auto, rezende2014stochastic, zhang2017split}, for time-series modelling~\cite{ghahramani2000variational, kulkarni2014variational, bayer2014learning, chung2015recurrent, fraccaro2016sequential}, and recently for RL~\cite{levine2018reinforcement, fellows2018virel, achiam2018variational}.

Our method builds on SRNN~\cite{fraccaro2016sequential}, which was originally presented for unstructured continuous latent variables, extending it to hierarchical control policies.
Our method also extends PHP~\cite{fox2018parametrized}, by training multi-level hierarchies jointly, rather than level-by-level.

\section{Preliminaries}
\slab{prelim}

\subsection{Imitation Learning as Stochastic Inference}
\slab{il}

We model an agent's interaction with its environment as a Partially Observable Markov Decision Process (POMDP).
At time $t$, the environment is in state $s_t \in \C{S}$, and emits an observation $o_t \in \C{O}$.
Upon seeing the observation, the agent makes a stochastic choice of an update for its internal memory state to $m_t \in \C{M}$, and of an action $a_t \in \C{A}$ (discrete, in this work), according to a policy $p_\theta(m_t, a_t | m_{t-1}, o_t)$.
The environment then makes a stochastic transition to the next state and observation $p(s_{t+1}, o_{t+1} | s_t, a_t)$.

We consider the imitation learning (IL) setting in which a teacher provides a set of demonstrations $\C{D} = \{x_j\}$.
Each demonstration $x = o_0, a_0, o_1, a_1, \ldots, o_T, a_T$ is an observation--action trace induced by the execution of a teacher policy in the environment.
Here $a_T$ is the first occurrence in the sequence of a \emph{termination} action $\nul$.
The learning problem is to find the parameter $\theta$ of a policy $p_\theta$ that gives high likelihood to the observed data, i.e. to maximize $\log p_\theta(\C{D}) = \sum_j \log p_\theta(x_j)$.
Denoting by $z = m_0, \ldots, m_T$ the latent sequence of memory states, we have
\eqn{
\log p_\theta(x) &= \log \sum_z p_\theta(z, x) \elab{opt} \\
&= \log \sum_z \prod_{t=0}^T p_\theta(m_t, a_t | m_{t-1}, o_t) + \const, \nn
}
with the constant in $\theta$ incorporating the log-probability of the environment steps $p(s_{t+1}, o_{t+1} | s_t, a_t)$.

When the memory update is non-deterministic, the support of $z$ can have size exponential in the horizon $T$, which prevents direct optimization of~\eref{opt}.
Many existing approaches either have no memory state, allowing only reactive policies $p_\theta(a_t | o_t)$, or more generally have deterministic memory updates $m_t = g_\theta(m_{t-1}, o_t)$, and by extension $z = g_\theta(x)$, e.g. using recurrent neural networks (RNNs).
This simplifies~\eref{opt} to $\sum_t \log p_\theta(a_t | m_{t-1}, o_t)$, which permits direct optimization, e.g. using back-propagation through time.

Unfortunately, when the internal memory process $z$ consists of discrete variables, such as the choice of procedures to invoke in a hierarchical structure, it is no longer representable as a sequence of differentiable memory-update functions.
To allow gradient-based optimization methods, we relax the RNN to be stochastic, and optimize in the space of stochastic policies.
As mentioned above, this in turn makes it infeasible to enumerate the space of latent memory trajectories.

Instead, we optimize the stochastic RNN with variational inference~\cite{fraccaro2016sequential}.
As detailed in~\sref{vi}, we train an \emph{inference model} $q_\phi(z | x)$ from which we can sample $z$ rather than exhaust its potentially large support.
The sampled $z$ then guides the training of the \emph{generative model} $p_\theta(z, x)$, which in this work has the PHP structure described in the next section.

\subsection{Parametrized Hierarchical Procedures}
\slab{php}

Structure in the policy $p_\theta$ is an inductive bias that can facilitate sample-efficient generalization from the demonstrated behavior to execution on unseen states.
In this work, we use the hierarchical control framework of parametrized hierarchical procedures (PHP)~\cite{fox2018parametrized}, in which a control policy is modelled as a set of procedures $\C{H}$.
Each procedure is tasked with performing a specific behavior, which it does by breaking the task down into a sequence of simpler sub-tasks, and calling a sequence of sub-procedures and control actions to perform these sub-tasks.

\begin{algorithm}[t]
\caption{PHP Step}
\alab{hier}
\begin{algorithmic}
\REQUIRE top stack frame $(h_i, \tau_i)$, PHP step selector $u_i$, current observation $o_t$
\IF{$u_i = \nul$ \COMMENT{$h_i$ terminates}}
    \STATE{Pop the top stack frame}
    \STATE{If stack is empty, terminate episode with action $a_t = \nul$}
\ELSE
    \STATE{Increment the top step counter $\tau_i$}
    \IF{$u_i \in \C{A}$ \COMMENT{$h_i$ takes elementary action $u_i$}}
        \STATE{Take action $a_t = u_i$ in the environment}
    \ELSIF{$u_i \in \C{H}$ \COMMENT{$h_i$ calls sub-procedure $u_i$}}
        \STATE{Push $(u_i, 0)$ onto the stack}
    \ENDIF
\ENDIF
\end{algorithmic}
\end{algorithm}

Similarly to computer programs, the hierarchical controller's execution is managed by a call-stack, onto which sub-procedures are pushed when called, and from which they are popped when they terminate (\aref{hier}).
Each \emph{frame} $(h, \tau)$ in the stack consists of the one-hot encoded procedure identifier $h \in \C{H}$ and the counter $\tau \in \F{N}$ of steps that the procedure executed so far.
Initially, the stack consists of a single frame, $(h_0, 0)$, where $h_0$ is a fixed \emph{root} procedure.
Let step $i$ of the execution occur in time $t$, and let the top stack frame be $(h_i, \tau_i)$.
Then procedure $h_i$ takes a stochastic step with distribution $p_\theta(u_i | h_i, \tau_i, o_t)$, where $u_i \in \C{H}\cup\C{A}\cup\{\nul\}$ is either a sub-procedure, a control action, or the termination indicator $\nul$.
Respectively in these three cases: (1) if $u_i \in \C{H}$, the frame $(u_i, 0)$ is pushed into the stack, calling the sub-procedure $h_{i+1} = u_i$; (2) if $u_i \in \C{A}$, the action $a_t = u_i$ is taken in the environment, and $h_i$ remains the active procedure; or (3) if $u_i = \nul$, then the frame $(h_i, \tau_i)$ is popped from the stack, and $h_i$'s caller (now at the top) resumes control.
In either of the cases (1) or (2), $h_i$'s step counter $\tau_i$ is incremented by 1.
The episode ends with $a_T = \nul$ when the root terminates.

\begin{figure}[t]
\centering
\includegraphics[width=0.285\columnwidth]{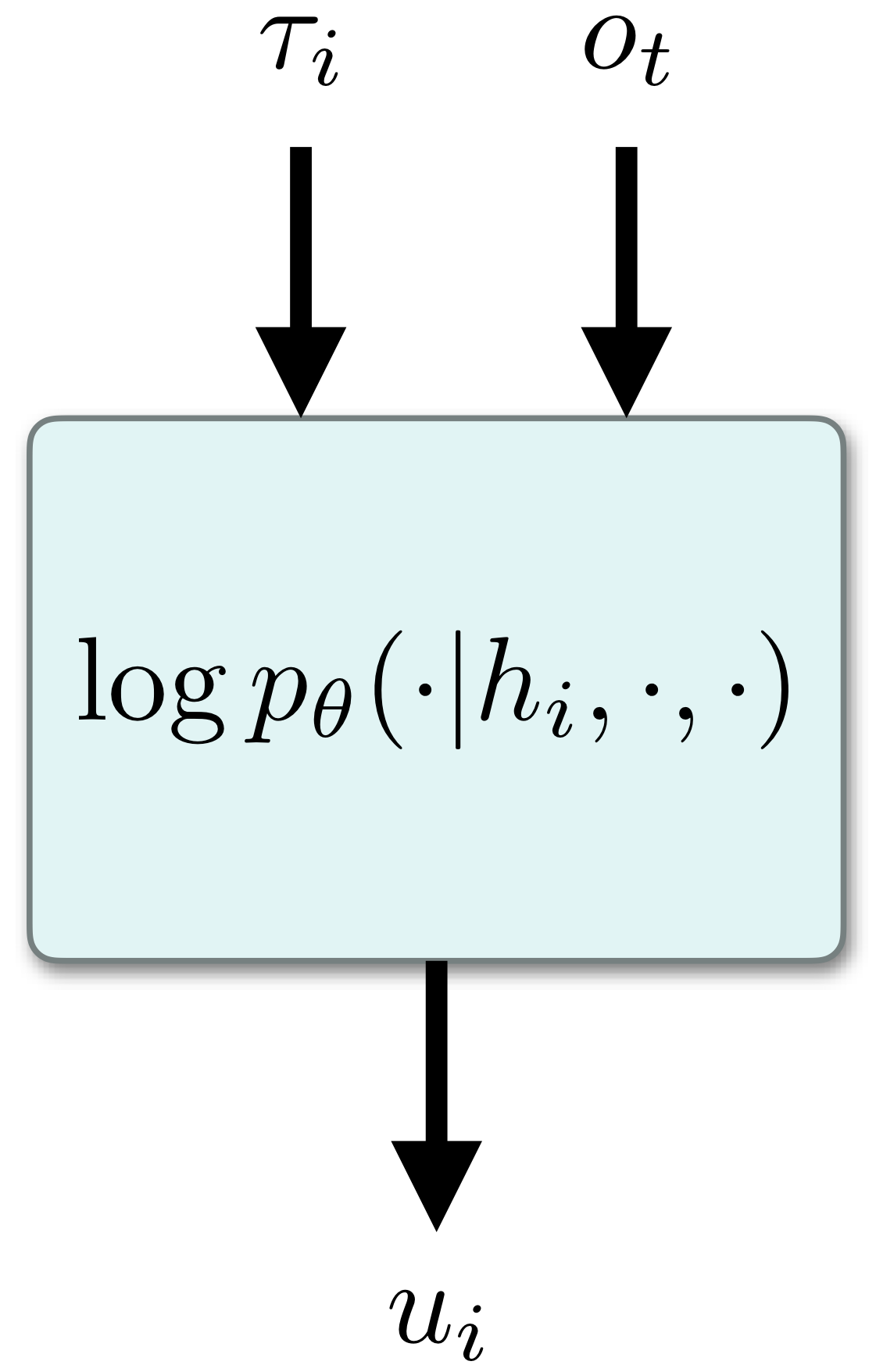}
\caption{Illustration of a parametrized hierarchical procedure. A control program is a hierarchical policy assembled from a set of procedures, one for each $h \in \C{H}$. A procedure can call another as sub-procedure by outputting the callee's identifier, or otherwise output the identifier of a control action to take it in the environment, or the indicator $\nul$ to terminate and return to its caller.}
\flab{gen}
\end{figure}

For each $h$, the PHP $p_\theta(u | h, \tau, o)$ is represented by a neural network (\fref{gen}), and jointly they all induce the stochastic process $z = \{u_i\}$.
Note that the observation $o_t$ is not stored on the stack, and that the time $t$ advances only in PHP steps $i$ that take control actions $a_t = u_i \in \C{A}$.

Each PHP can in principle call any other PHP, including itself recursively.
In practice, it is useful to define a \emph{call-graph} over procedures, with edges from each PHP to those it can call as sub-procedures.
The call-graph is a way for users to specify any available prior knowledge on the desired program structure.
In our experiments, unless otherwise specified, we restrict call-graphs to be full 5-ary trees.

\subsection{Variational Inference}

Since we cannot directly optimize the objective~\eref{opt}, we will optimize a proxy, namely the evidence lower bound (ELBO).
For any distribution $q_\phi(z | x)$, we have
\eqn{
\log p_\theta(x) \ge \E_{z | x \sim q_\phi}\left[\log \frac{p_\theta(z, x)}{q_\phi(z | x)}\right], \elab{elbo}
}
where the bounding gap is $\D[q_\phi(z | x) \| p_\theta(z | x)]$, the Kullback--Leibler (KL) divergence of the inference model $q_\phi(z | x)$ from the true, computationally infeasible posterior of the generative model $p_\theta(z | x)$.
Maximizing the lower bound on the right-hand side of~\eref{elbo} over $\theta$ and $\phi$ is therefore trading off maximizing our log-likelihood objective on the left-hand side, with minimizing the gap.

We optimize the ELBO by sampling a batch of traces from the dataset, for each trace $x$ sampling $z | x \sim q_\phi$, and computing the \emph{score}
\eqn{
f_{\theta, \phi}(z, x) = \log \frac{p_\theta(z, x)}{q_\phi(z | x)}. \elab{score}
}
The gradient of $f$ with respect to $\theta$ is an unbiased estimator for that of the ELBO, but the same is not true with respect to $\phi$ due to the dependence of the sampling distribution on $\phi$.
Since $z$ is discrete and non-reparametrizable, we use the score-function trick, estimating ELBO's gradient as
\eqn{
\nabla f(z, x) + f(z, x) \nabla \log q_\phi(z | x). \elab{grad}
}

\section{Hierarchical Variational Imitation Learning}
\slab{vi}

\subsection{Variational Inference of Control Programs}

\begin{algorithm}[t]
\caption{HVIL}
\alab{vi}
\begin{algorithmic}
\STATE Initialize $p_\theta$ and $q_\phi$
\LOOP
    \STATE Sample batch of demonstration traces from $\C{D}$
    \STATE Initialize $g = 0$
    \FOR{each trace $x$}
        \STATE Get posterior context $\{b_t\}$ from bidirectional RNN
        \STATE Initialize $t = 0$, $\ell = 0$
        \FOR{each PHP step $i$}
            \STATE Let $(h_i, \tau_i)$ be the top stack frame
            \STATE $D \gets \D[q_\phi(u_i | h_i, \tau_i, b_t) \| p_\theta(u_i | h_i, \tau_i, o_t)]$
            \STATE $g \gets g + \nabla_{\theta, \phi} D + D \ell$
            \STATE Sample $u_i | h_i, \tau_i, b_t \sim q_\phi$
            \STATE $\ell \gets \ell + \nabla \log q_\phi(u_i | h_i, \tau_i, b_t)$
            \STATE Operate the stack using~\aref{hier}
            \STATE If $u_i \in \C{A}$, set $t \gets t + 1$
        \ENDFOR
    \ENDFOR
    \STATE Take gradient descent step $g$
\ENDLOOP
\end{algorithmic}
\end{algorithm}

Hierarchical Variational Imitation Learning (HVIL), summarized in~\aref{vi}, uses variational inference to learn a PHP from demonstrations.
Following the notation of~\sref{prelim}, we need to be able to compute $\log p_\theta(z, x)$ and $\log q_\phi(z | x)$, and to sample from $q_\phi(z | x)$.
For the control structure defined in~\sref{php}, we have
\eq{
\log p_\theta(z, x) = \sum_t \sum_{i \in \C{I}_t} \log p_\theta(u_i | h_i, \tau_i, o_t) + \const,
}
where $\C{I}_t$ are the PHP steps taken in time $t$, and the constant is the same as in~\eref{opt}.
Similarly, it is convenient to choose for $q_\phi$ the structure
\eq{
\log q_\phi(z | x) = \sum_t \sum_{i \in \C{I}_t} \log q_\phi(u_i | h_i, \tau_i, x).
}

\begin{figure}[t]
\centering
\includegraphics[width=0.8\columnwidth]{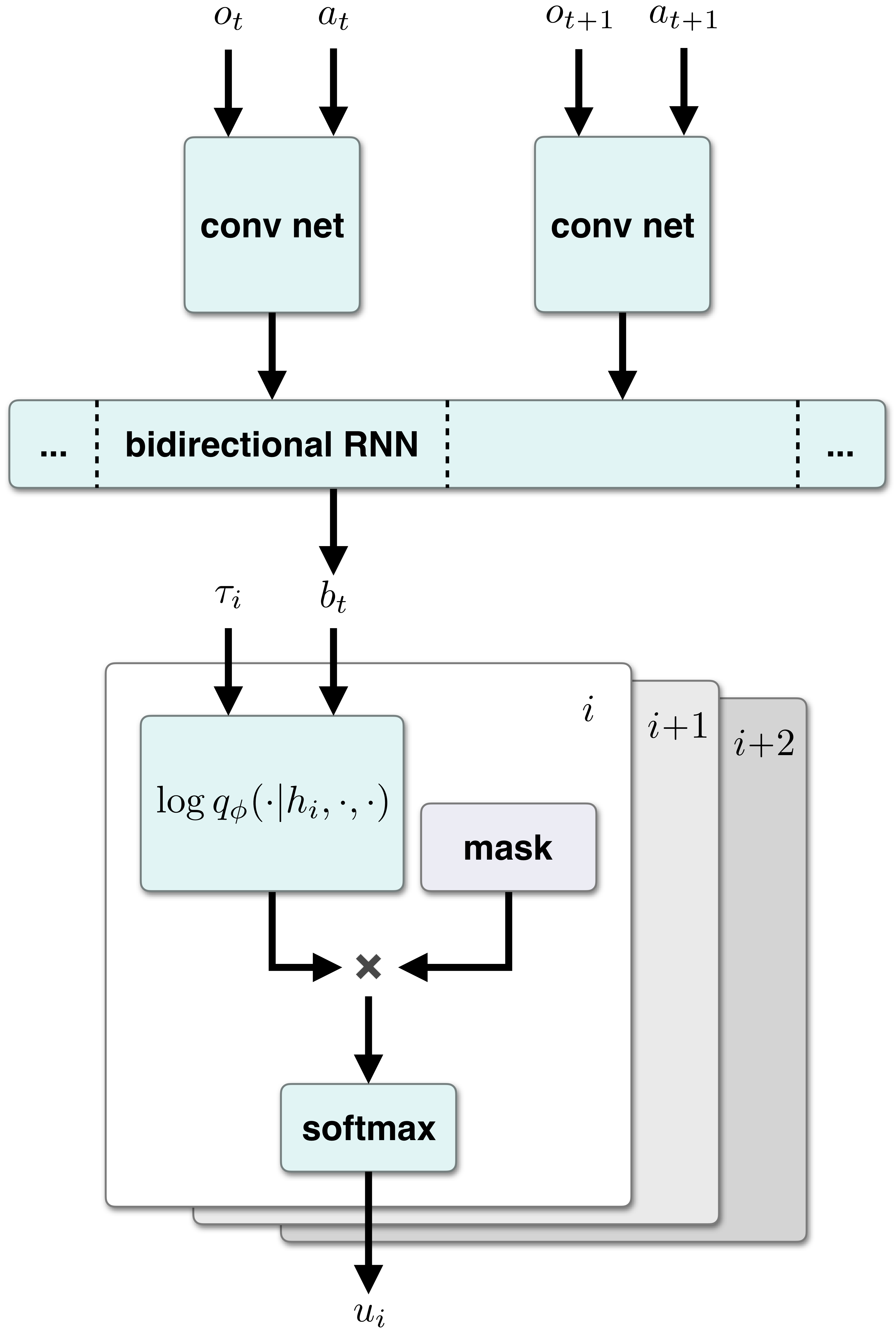}
\caption{Architecture of an inference procedure. The demonstration trace $x = \{o_t, a_t\}$ is processed by a bidirectional RNN to generate a posterior context $b_t$, which feeds into a network computing a distribution $\log q_\phi(u_i | h_i, \tau_i, b_t)$ over the PHP step (sub-procedure call, control action, or termination). The $u_i$ logits are masked to preclude PHP steps that are inconsistent with the trace $x$.}
\flab{inf}
\end{figure}

This structure suggests parametrizing $q_\phi$ using a second copy of the set of procedures used by $p_\theta$.
Such \emph{inference procedures}, illustrated in~\fref{inf}, would operate similarly to the original \emph{generative procedures}, with two differences.
First, the distribution $q_\phi$ is conditioned on the entire trace $x$, both its past and future, which implies that inference procedures should get $x$ as input.
Following the SRNN architecture~\cite{fraccaro2016sequential}, we process the trace $x$ using a deterministic bidirectional RNN $b_\phi(x)$, to obtain a time-dependent \emph{posterior context} $b_t$.
We use $b_t$ as input to inference procedures in the same way that we input $o_t$ to generative procedures, yielding $q_\phi(u_i | h_i, \tau_i, b_t)$.
Note that gradients of $q_\phi$ and $p_\theta$ are not back-propagated through the discrete variables $h_i$, $\tau_i$, or $u_i$, but are back-propagated to the bidirectional RNN through $b_t$.

The second difference from generative procedures is that inference procedures are constrained to make choices that includes all the correct actions of the demonstrated trace, i.e. having $u_i = a_t$ for the last $i$ in $\C{I}_t$ for all $t$.
Otherwise, $q_\phi$ has support for inconsistent values of $z$ for which $p_\theta(z, x) = 0$, leading to an undefined ELBO.
To achieve consistency, we mask the output of the active inference procedure to set the logit of any inconsistent $u_i$ to $-\infty$, before normalizing $q_\phi$ with softmax.
If any prior information is available in the form of a call-graph, restricting which sub-procedures each procedure can call, then we ensure that inference procedures only take paths in this graph that can lead to the known action $a_t$.
In particular, we preclude any control action selection other than $a_t$.

\subsection{Variance Reduction}
\slab{tech}

\paragraph{Analytic KL computation.}

Computing the ELBO~\eref{elbo} ``analytically'', with an explicit sum over all values of $z$, often has better convergence properties than sampling $z | x \sim q_\phi$~\cite{kingma2013auto}.
This is usually attributed to the high variance of the score~\eref{score}, particularly when $q_\phi(z | x)$ is a poor approximation of the true posterior $p_\theta(z | x)$.
In our experiments, we found that analytic KL was indeed necessary, but for a different reason: the ELBO gradient estimator~\eref{grad} becomes effectively biased as $q_\phi(z | x)$ tend to 0 for some values of $z$ that are supported by $p_\theta$.
This occurs frequently in our setting when $q_\phi$ is successful in using future information in $x$, unavailable to $p_\theta$, to rule out ``incorrect'' latent structure.

We prove the existence of this effect in a simple binary example, where $q(z | x)$ is $\epsilon$ for $z = 0$, and $1 - \epsilon$ for $z = 1$.
We note that, as $\epsilon \to 0$, samples of $z$ of any practical size eventually cease to include instances of $z = 0$.
The score gradient estimator (to maximize) then becomes
\eq{
\MoveEqLeft \dif_\epsilon f + f \dif_\epsilon \log q(z | x) \\
&= -\dif_\epsilon \log (1 - \epsilon) + \frac {\log p(1, x)}{\log (1 - \epsilon)} \dif_\epsilon \log (1 - \epsilon) \\
&\approx 1 - \frac{\log p(1, x)}{1 - \epsilon} > 0.
}
Intriguingly, even as $q$ approaches the correct distribution and samples correct values of $z$, the gradient indicates, incorrectly, that the objective would be improved by increasing $\epsilon$.
This would prevent the convergence of $q$ to a correct deterministic distribution.

To compute the ELBO~\eref{elbo} analytically, we must factorize the exponentially large support of $q_\phi$ by breaking the ELBO down into individual PHP steps, and then compute the KL of each step with an exhaustive sum.
We define
\eq{
D_{t, i}(u_{<i}, x) = \D[q_\phi(u_i | u_{<i}, x) \| p_\theta(u_i | u_{<i}, o_t)],
}
such that the ELBO is
\eqn{
-\sum_t \sum_{i \in \C{I}_t} \E_{u_{<i} | x \sim q_\phi} [D_{t, i}(u_{<i}, x)]. \elab{factor}
}
For each PHP step $i$, we sample $u_{<i} = u_0, \ldots, u_{i-1}$ by extending $u_{<i-1}$ from the previous step, and compute the one-step KL $D_{t, i}(u_{<i}, x)$ analytically.
This process is summarized in~\aref{vi}.

\paragraph{Rao--Blackwellization of analytic-KL score gradient estimators.}

Rao--Blackwellization is the reduction of the variance of an estimator by taking its expectation with respect to a sufficient statistic.
The score gradient estimator~\eref{grad} can be Rao--Blackwellized by including in $\nabla \log q_\phi (z | x)$ only factors of $z$ that are used in computing $f$~\cite{schulman2015gradient}.

We leverage our ELBO factorization~\eref{factor} to propose a Rao--Blackwellization of a score gradient estimator that is computed with analytic KL:
\eq{
\sum_t \sum_{i \in \C{I}_t} (\nabla D_i(u_{<i}, x) + D_i(u_{<i}, x) \nabla \log q_\phi(u_{<i} | x)).
}

\subsection{Leveraging Acausal Information}

In this section we identify a source of information in demonstration traces that, to our knowledge, has previously been unused.
We show that HVIL can extract this information and leverage it to improve data efficiency.
We demonstrate this principle with an illustrative example.

Consider a conditional generative model $p(y | x)$.
There may exist a variable $z$, such that knowing $z$ at training time breaks down the problem into two easier learning problems: of estimating $p(z | x)$ and $p(y | x, z)$.
For example, let $x$ be an image, $z$ a label in one of 10 classes, and $y$ an arbitrary partition of the classes into two categories.
Such $y$ is less informative than $z$ about important image features, and its dependence on those features is higher-order than that of $z$, leading to higher sample complexity when $z$ is latent.

Similar arrangements can occur naturally in imitation learning problems.
Consider a robot learning to manipulate objects on a table.
The robot perceives an image $x$ of an object, and decides on an action $y$, e.g. pick or push.
Later, the robot reaches for the object to pick or push it, and can take a closer and clearer image, in which the object's class $z$ is obvious.
Since $z$ is not causally linked to $y$, most imitation learning algorithms do not extract from $z$ any information useful for training the policy $p(y | x)$.

To illustrate this phenomenon, we define a simple POMDP in which $o_0$ is an MNIST image and $o_1$ is a one-hot encoding of the digit in that image.
The teacher selects a binary action $a_0$ that equals the digit's parity with probability $p$ varying from 0.6 to 1, and terminates on $a_1 = \nul$.
An RNN trained with back-propagation through time to minimize the cross-entropy from teacher demonstrations has no gradient in time $t = 1$, because $a_1$ is constant.
Although $o_1$ is very informative, it is never used in training time because it cannot be used to decide $a_0$ in execution time.

We generate datasets of between 100 and 6400 traces, and train three models: (1) a 64-unit LSTM, fed by a standard ConvNet\footnote{The architecture for the MNIST ConvNet is taken from \url{https://github.com/pytorch/examples/blob/master/mnist/main.py}}; (2) a similar 2-layer LSTM; and (3) a 2-level PHP with one root and 5 sub-procedures.
The generative procedures are the same standard ConvNet, and the inference procedures use a 32-unit bidirectional LSTM, fed by a ConvNet and feeding into a multi-layer perceptron (MLP) with a 64-unit hidden layer.

\begin{figure}[t]
\centering
\subfigure[]{
    \includegraphics[width=0.45\columnwidth]{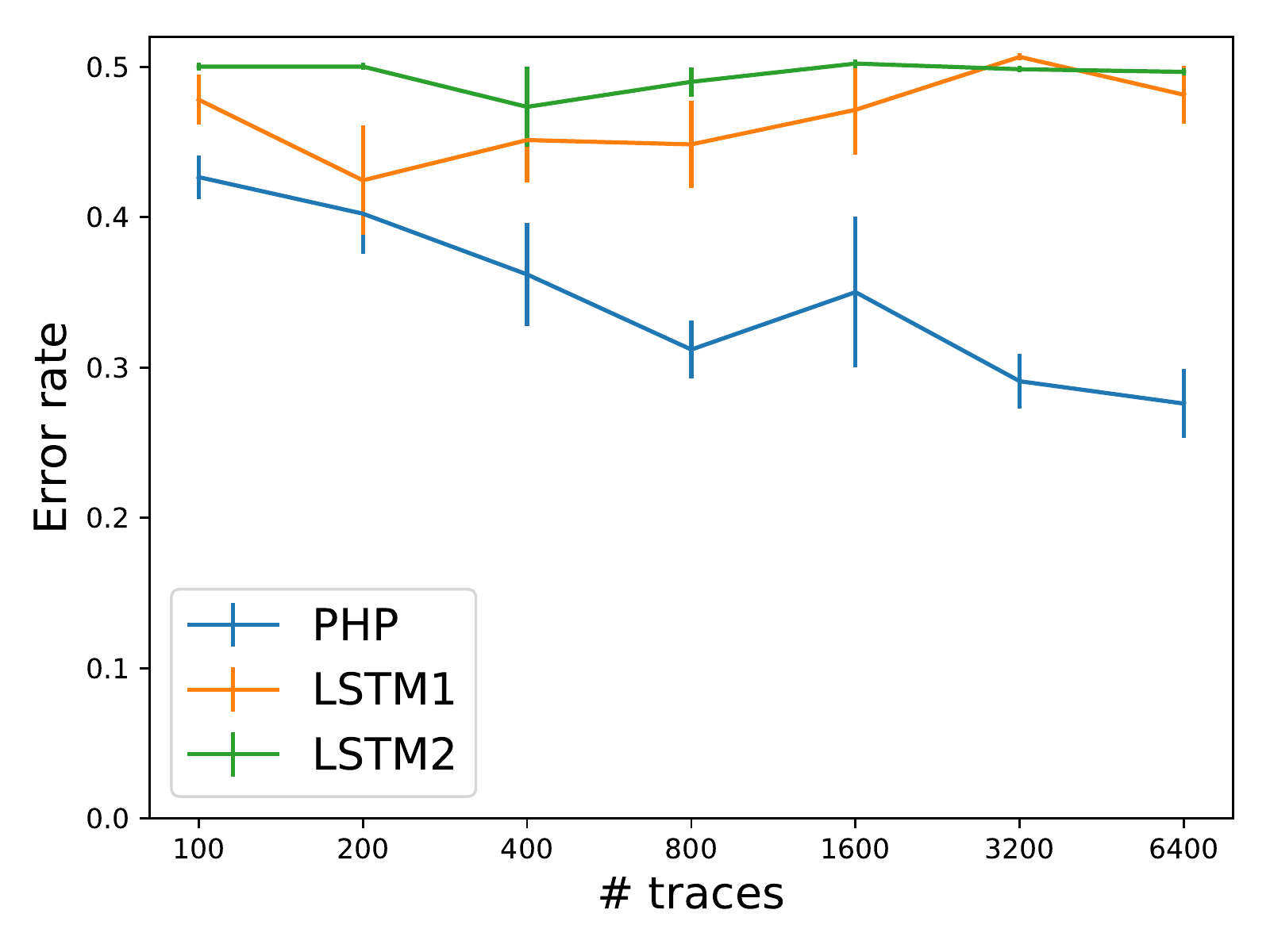}
}
\subfigure[]{
    \includegraphics[width=0.45\columnwidth]{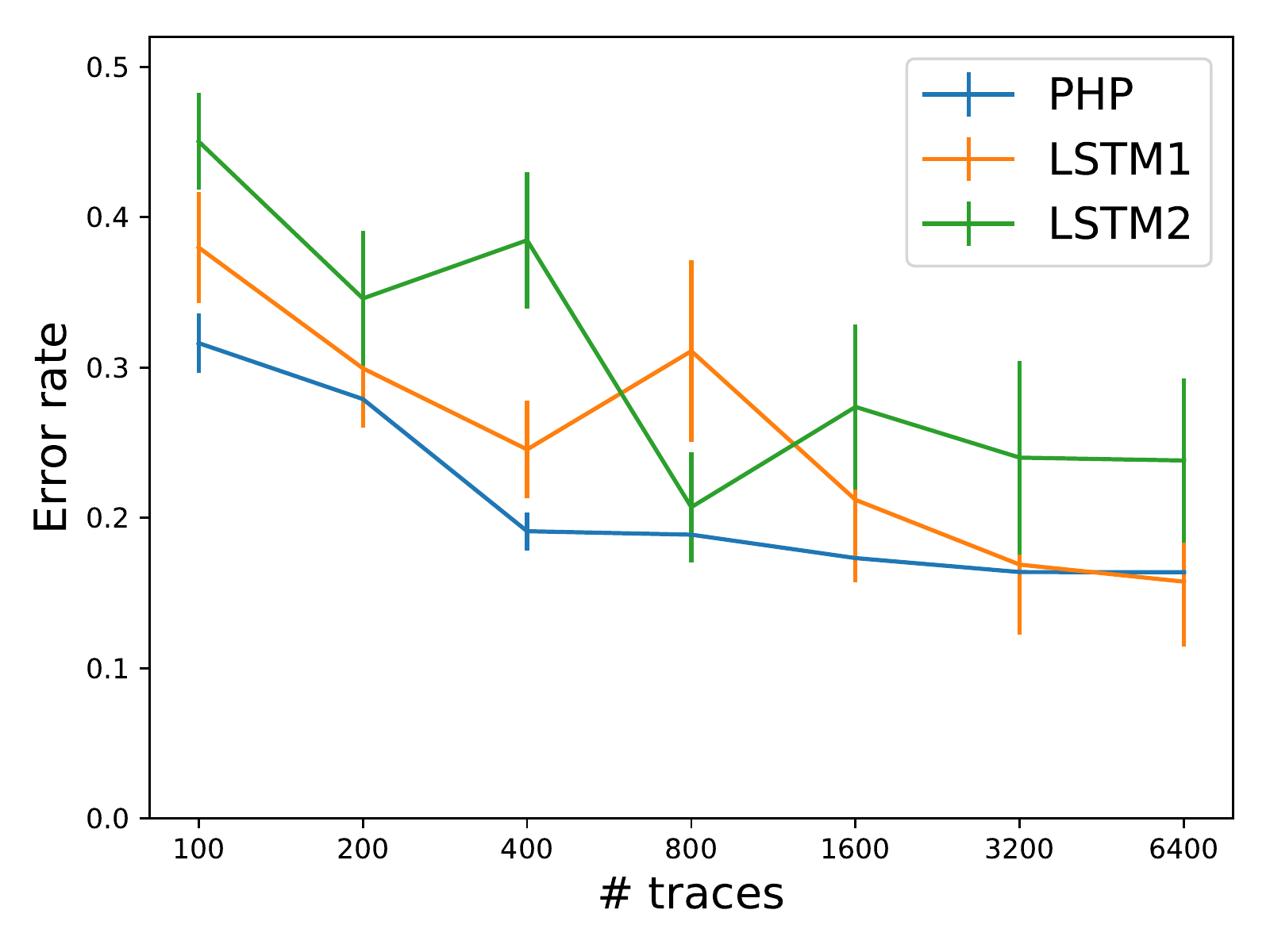}
}
\subfigure[]{
    \includegraphics[width=0.45\columnwidth]{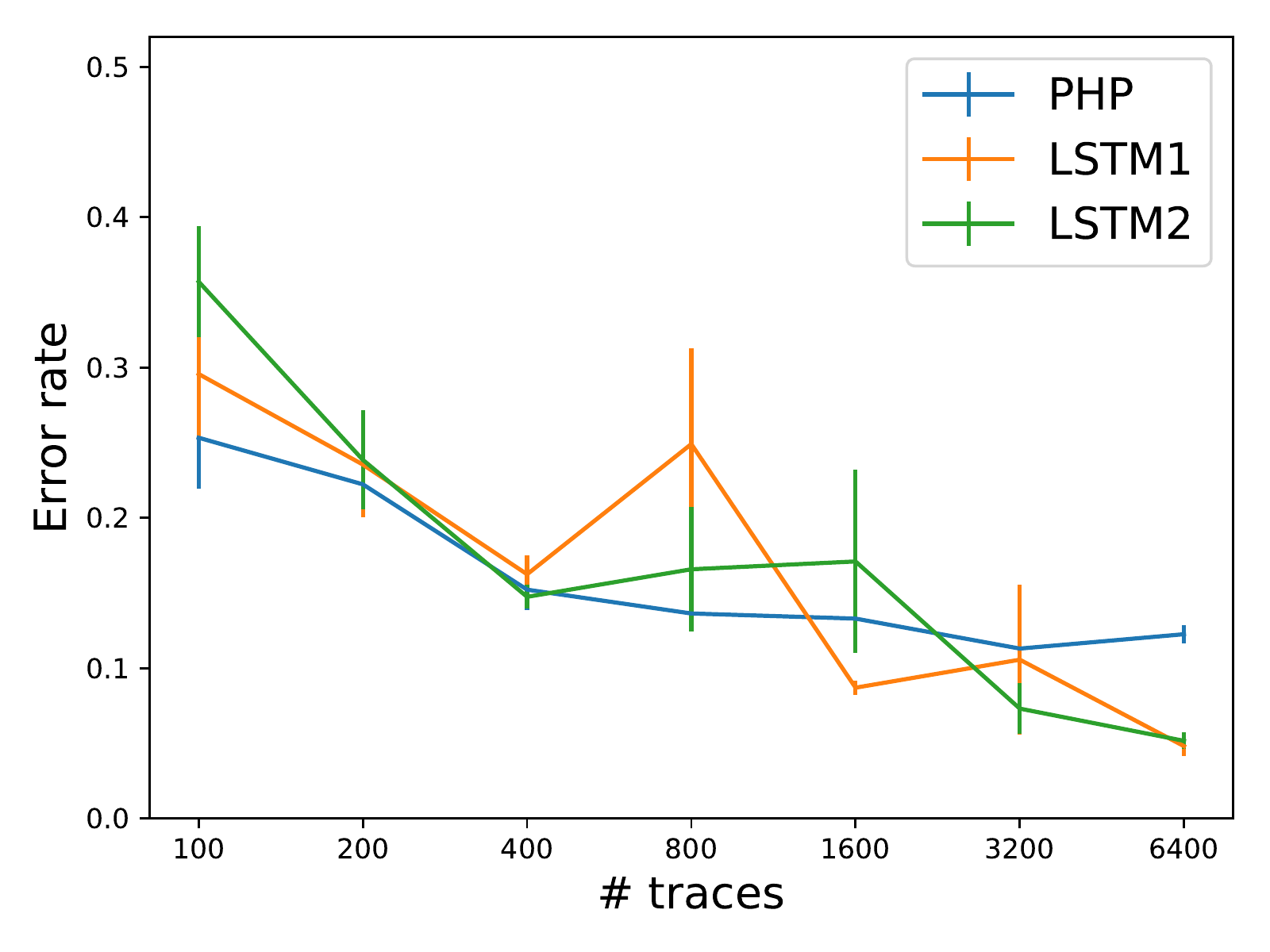}
}
\subfigure[]{
    \includegraphics[width=0.45\columnwidth]{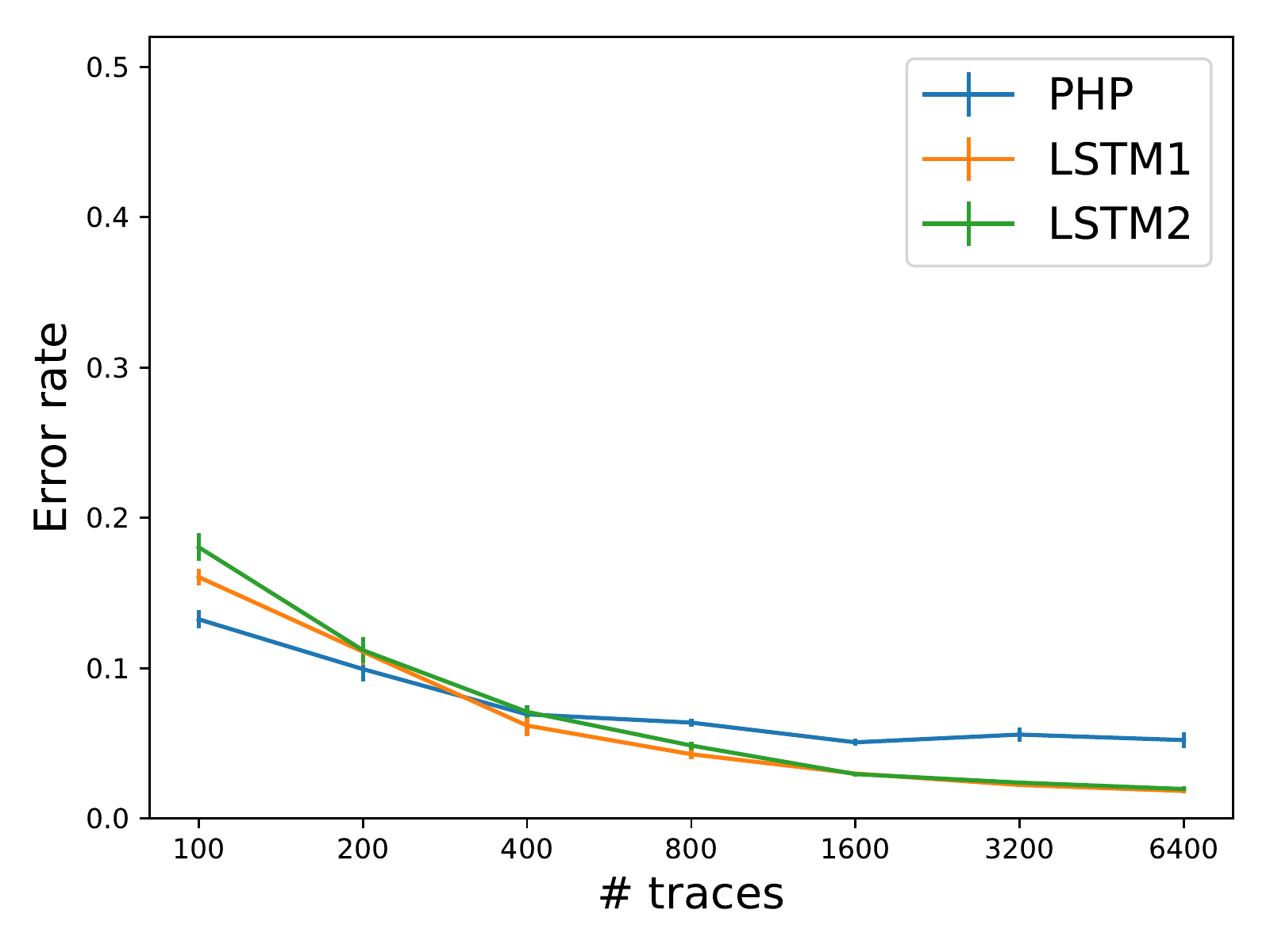}
}
\caption{Error rate of imitating a teacher outputting the parity of MNIST digits and terminating. In training time, the teacher demonstrates the correct parity with probability 0.6, 0.7, 0.8, and 1, respectively in (a)--(d). The number of demonstration traces ranges from 100 to 6400 on a log scale. The error rate is averaged over 10 runs. Our method, 2-level PHP trained with HVIL, outperforms both 1- and 2-layer LSTM baselines in the low-data, high-noise regime. This indicates that HVIL leverages acausal information in the data to achieve improved data efficiency.}
\flab{acausal}
\end{figure}

The results, summarized in~\fref{acausal}, show a clear benefit to PHP trained with HVIL in the low-data, high-noise regime.
Even without noise, our method outperforms LSTM when only 100 demonstrations are given.
This suggests that the inference procedures were indeed able to infer latent structure $z$ that helps guides training of the generative procedures.

\section{Experiments}

\subsection{Bubble Sort}

\begin{figure}[t]
\centering
\includegraphics[width=0.8\columnwidth]{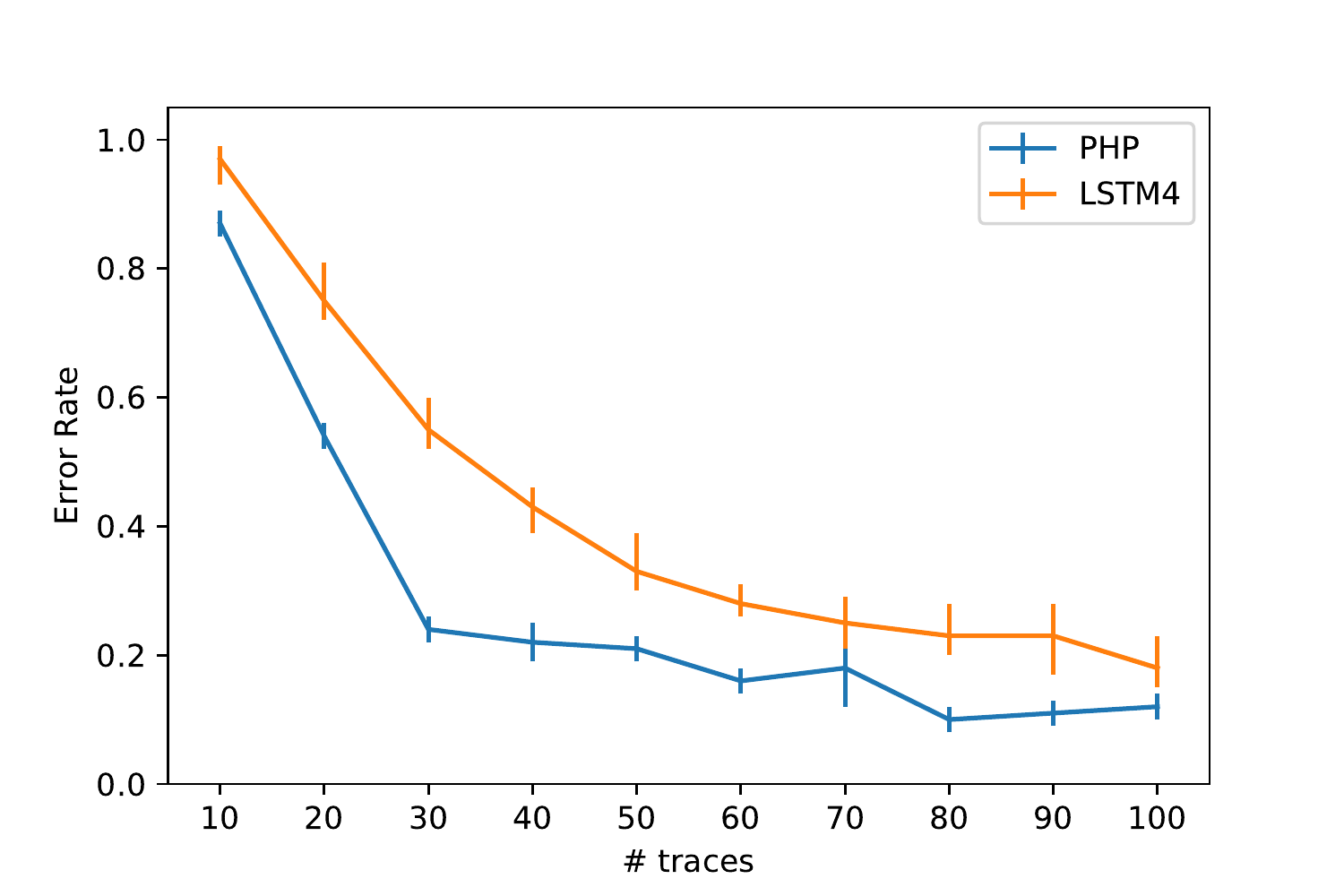}
\caption{Error rate of imitating a teacher to perform bubble sort. We compare training a 4-level 6-procedure PHP via HVIL, with end-to-end training of a 4-level LSTM. The error rate is the fraction of test traces with at least one wrong action, averaged over 3 runs. HVIL is able to extract more information from the same data by discovering a latent hierarchical structure that generalizes better, requiring less than half as much data to achieve a 24\% error rate.}
\flab{bubblesort}
\end{figure}

We train a hierarchical control policy to perform the bubble sort algorithm on a list of integers stored in external memory.
The environment is an emulator of a computation device with two memory pointers.
The actions can move either pointer left or right, or swap the memory values between the pointers.
Each observation consists of the values addressed by the two pointers as one-hot encodings, as well as indicators for each pointer addressing the first or last integer.

\paragraph{Experiment setup.}

A teacher generates traces of the execution of the bubble sort algorithm, on lists of up to 10 integers.
We use HVIL to learn a 4-level PHP, consisting of 6 procedures arranged in a partial binary tree.
Each generative procedure is represented by a MLP with a 100-unit hidden layer.
The inference procedures have the same architecture, applied to the output of a 32-unit bidirectional LSTM instead of the observation.

Every procedure in the hierarchy can take any control action, and in principle could perform the entire task.
To encourage diversity between the procedures and avoid mode collapse into a non-hierarchical solution, we add an entropy regularization term on the inference procedures with an initial weight of 1 and exponential decay of 0.7 every 5000 steps.

We compare this model to a 4-layer 64-unit LSTM, fed by a 64-unit MLP and feeding into another 64-unit MLP to which softmax is applied to select the action.

The optimization algorithm is Adam with weight decay $10^{-3}$ and PyTorch default hyperparameters.
Each batch consists of 10 traces.
Each model is trained for 100K steps from demonstration datasets of sizes varying between 10 and 100.

\paragraph{Results.}

\fref{bubblesort} summarizes the experiment results.
As expected, the error rate drops as larger datasets are used, and it drops faster for HVIL training of PHP than for the LSTM baseline.
These results suggest that HVIL is able to extract more information than end-to-end RNN training from the same amount of data, and to use it to execute the bubble sort algorithm more reliably.
With 30 demonstrations, HVIL already generalizes to reproduce the exact action sequence in 76\% of test traces, while RNN requires more than 70 demonstrations to achieve that level of performance.

\subsection{Karel}

Karel is an educational programming language~\cite{pattis1981karel} used in introductory classes.
In the Karel language, the programmer writes programs (complete with constructs like \texttt{if}, \texttt{ifElse}, and \texttt{while}) that produce sequences of actions for an agent (a robot named Karel) which lives in an $m \times n$ grid world.
Each cell in the grid can contain a wall, or up to 10 \emph{markers}.
The agent can occupy cells containing markers, but not walls.
The actions available to Karel are \texttt{move}, \texttt{turnLeft}, \texttt{turnRight} to advance along its current orientation or change the orientation, and \texttt{pickMarker}, \texttt{putMarker} to decrease or increase the number of markers in the agent's current cell.
The agent observes the values of \texttt{leftIsClear}, \texttt{rightIsClear}, \texttt{frontIsClear}, and \texttt{markersPresent}, which are the predicates (along with their negations) that can appear as the condition for \texttt{if}, \texttt{ifElse}, and \texttt{while}.

Karel has been used in past work on program synthesis and induction~\cite{devlin2017robustfill,bunel2018leveraging,shin2018improving,chen2018executionguided}.
In their setting, the method receives a number of input and output pairs that specify the semantics of an unknown program.
Then, the model should either learn to directly operate the robot to reproduce the correct final grid for unseen input grids (\emph{program induction}), or output a program in the Karel language that matches the unknown program (\emph{program synthesis}).
In our work, we perform program induction, represented as PHP, by imitating complete demonstration traces of a target Karel program.
That is, the trained PHP should performs the same actions as that Karel program on test inputs.

\paragraph{Experiment setup.}

We evaluate HVIL on the 7 Karel programs from \emph{Hour of Code} which contain loops (\url{https://bit.ly/karel-dataset}); they are listed in~\xref{programs}.
We prepend \texttt{turnRight} to each program, so that at least one step is taken before termination, as PHP requires.
In our experiments, we sought to compare the training data efficiency as well as generalization ability of our PHP training method, compared to an RNN baseline.

To generate our training and test data, we follow the methodology of~\citet{bunel2018leveraging}.
We randomly sample a set of 5 initial grid states.
For each grid, we: (1) randomly pick a world size between $2 \times 2$ and $16 \times 16$, excluding the outer walls; (2) sample whether each cell should contain walls or markers from a Bernoulli distribution (with its parameter drawn from a normal distribution); (3) for the marker-containing cells, sample a marker count from a geometric distribution; and (4) sample a location and orientation for the agent amongst the cells lacking walls.
After sampling 5 grids, we execute the program on them and check whether they execute without crashing (without attempting to move into a wall) and sufficiently exercise all paths through it (i.e. for each statement in the program, at least one grid causes it to be executed; each branch of if/while blocks are taken at least once).
If they do, we record the demonstrations using these initial grid states and add them to our dataset; otherwise, we reject them.
We repeat this process until we have a total of 10,000 demonstrations for each program.

We compare four different models. The first is an LSTM baseline containing two layers with 64 hidden units each, fed by a 64-unit MLP to process the observation and feeding into another 64-unit MLP to decide the action (where each MLP contains two fully-connected layers with a ReLU activation function in between).
The other three models are PHPs with full 5-ary trees of depth 1, 2, and 3 as call-graphs, each containing 1, 6, and 31 procedures respectively with a 256-unit hidden layer.
As the call-graphs are full 5-ary trees, they encode no domain-specific information about the structure of the learned program.

\paragraph{Data efficiency experiments.}

\begin{figure*}[t]
\centering
\hfill
\includegraphics[width=0.29\textwidth]{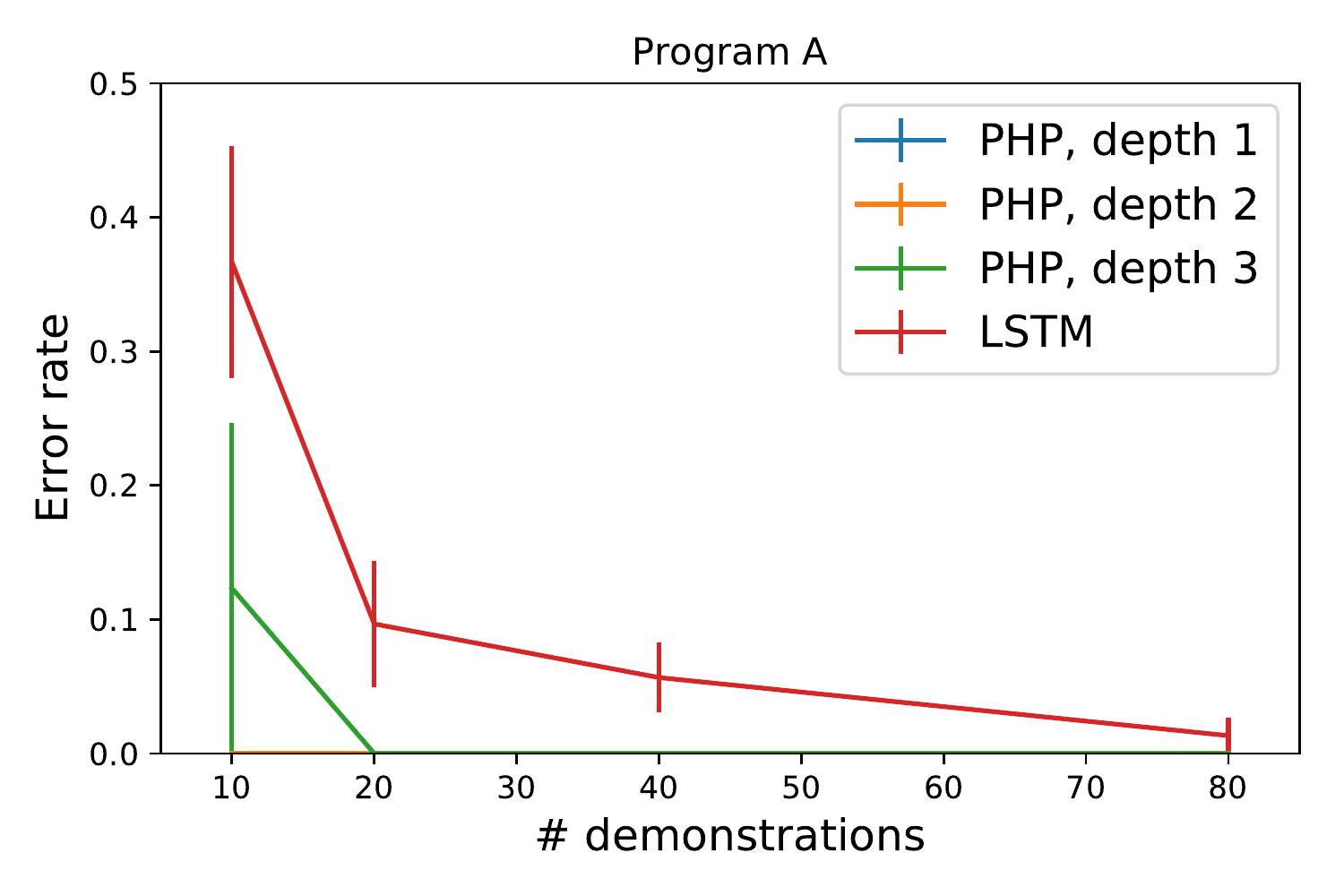}
\hfill
\includegraphics[width=0.29\textwidth]{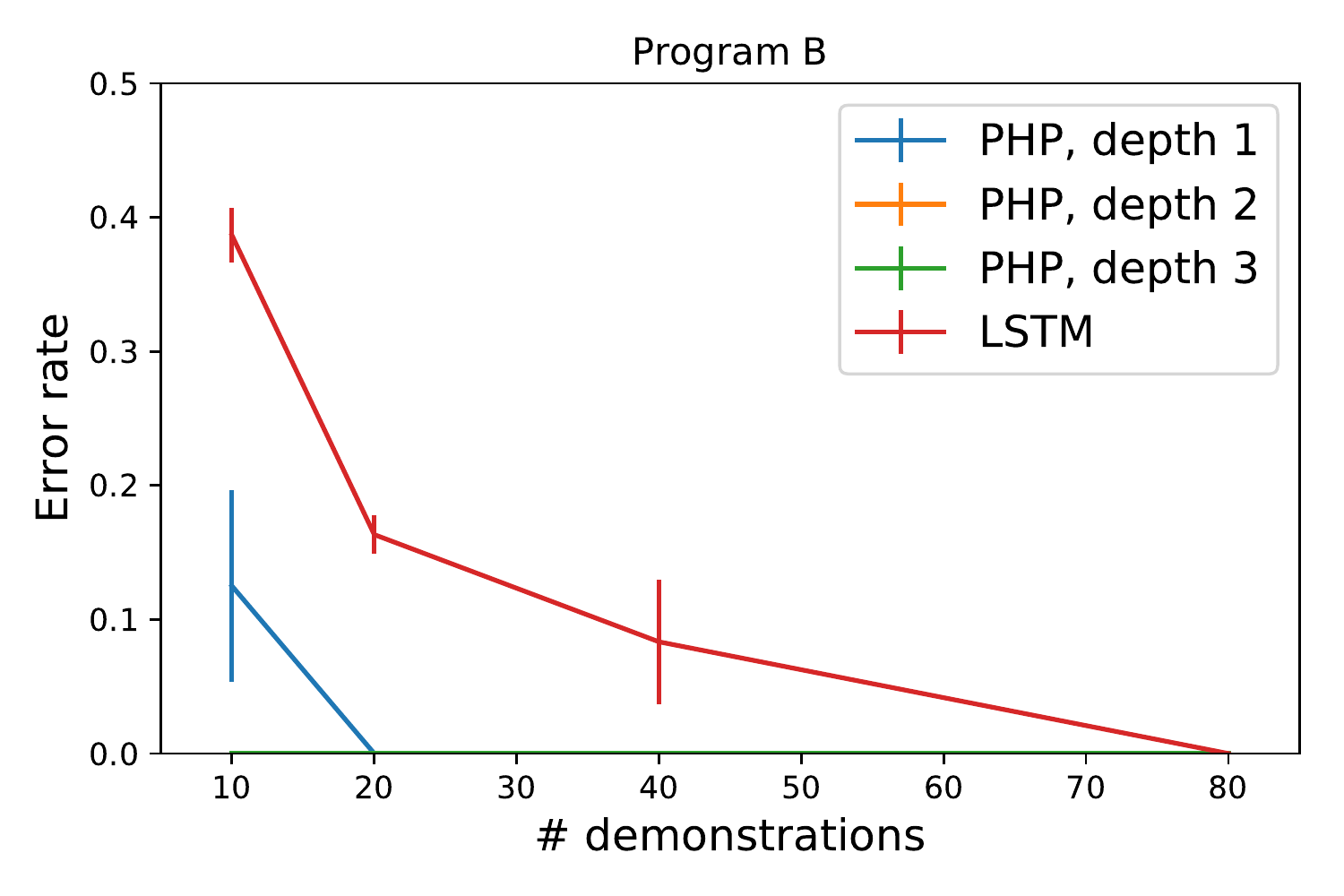}
\hfill
\includegraphics[width=0.29\textwidth]{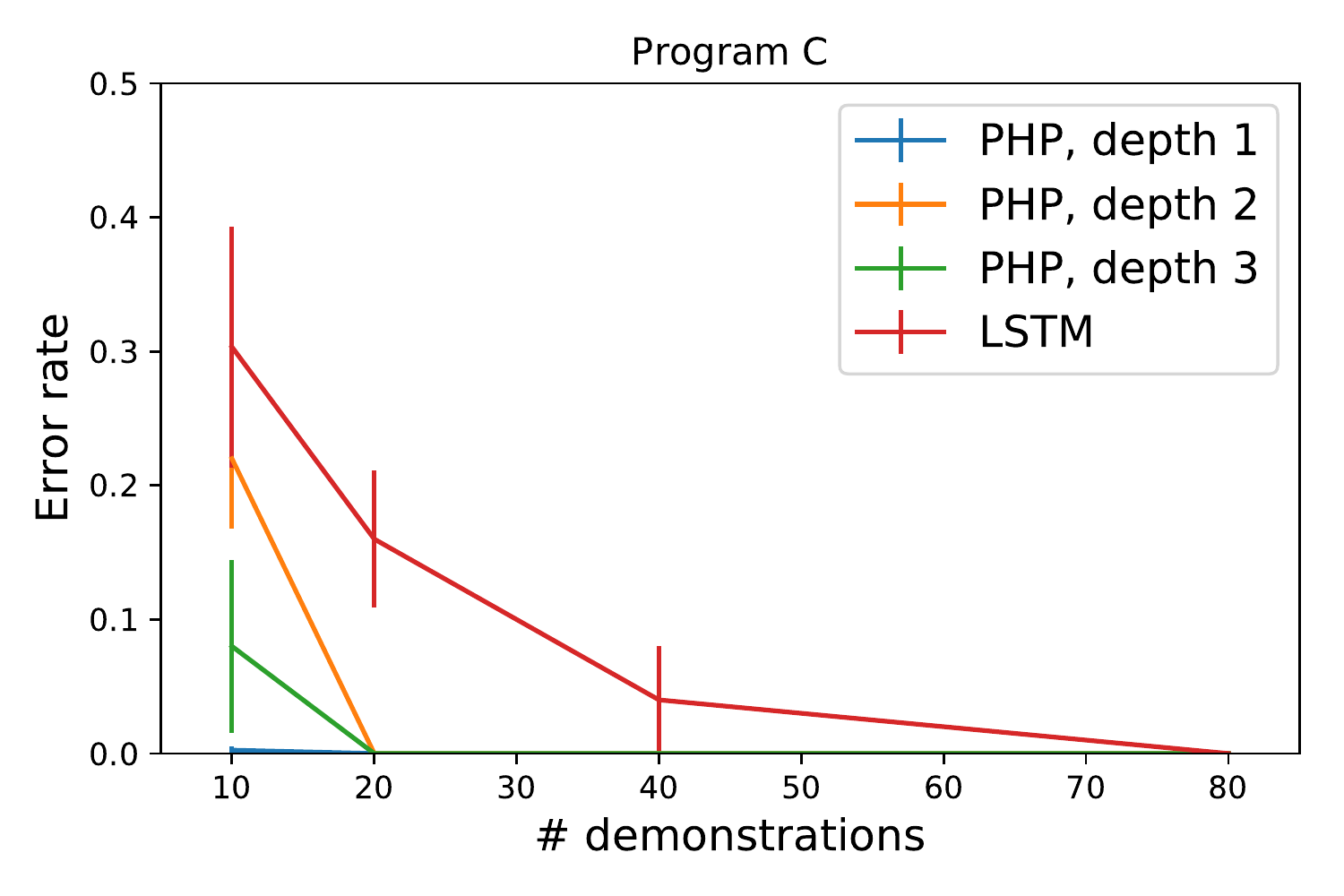}
\hfill
\caption{Error rate in imitation learning of Karel programs for datasets of training demonstrations of sizes 10 to 80.
We compare three different PHP arrangements, full 5-ary trees of depths 1 to 3, trained via HVIL, as well as an end-to-end trained LSTM baseline. Results for more Karel programs are included in~\xref{results}.}
\flab{karel-sample-efficiency}
\end{figure*}

In these experiments, we compare the performance of the four models on a test set containing 100 demonstrations, after training on 10, 20, 40, or 80 demonstrations randomly sampled from the remaining 9900 demonstrations.
\fref{karel-sample-efficiency} summarizes the results.

Overall, both kinds of models benefit from having a greater number of training demonstrations.
However, our results suggest that HVIL training of PHPs can again extract more information from the same amount of data compared to the LSTM baseline, as they generally achieve lower error.
The depth of the hierarchy has varying effect on the accuracy: in program C, the 3-level PHP performs best, but in program A worst.

\paragraph{Generalization experiments.}

\begin{figure*}[t]
\centering
\hfill
\includegraphics[width=0.29\textwidth]{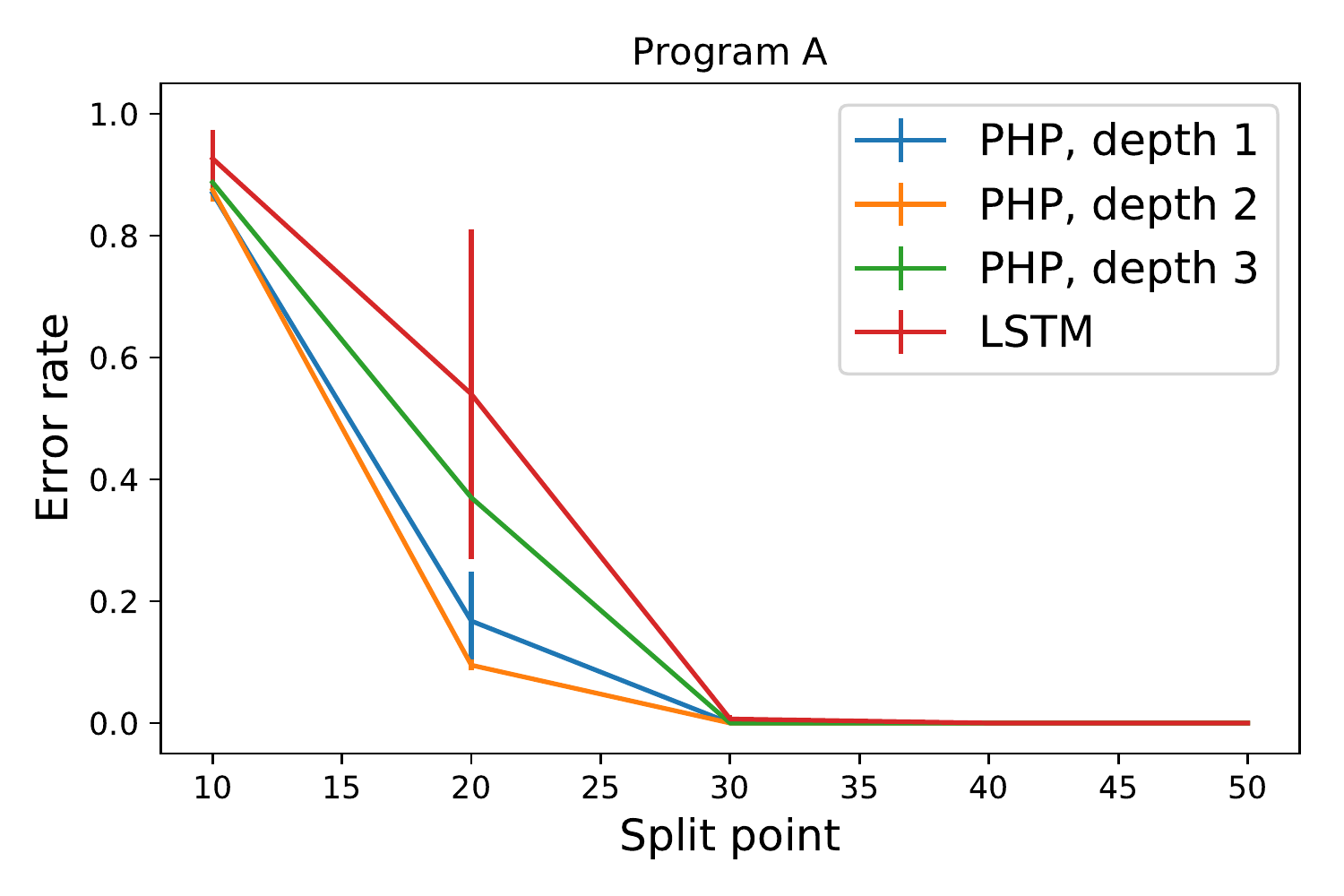}
\hfill
\includegraphics[width=0.29\textwidth]{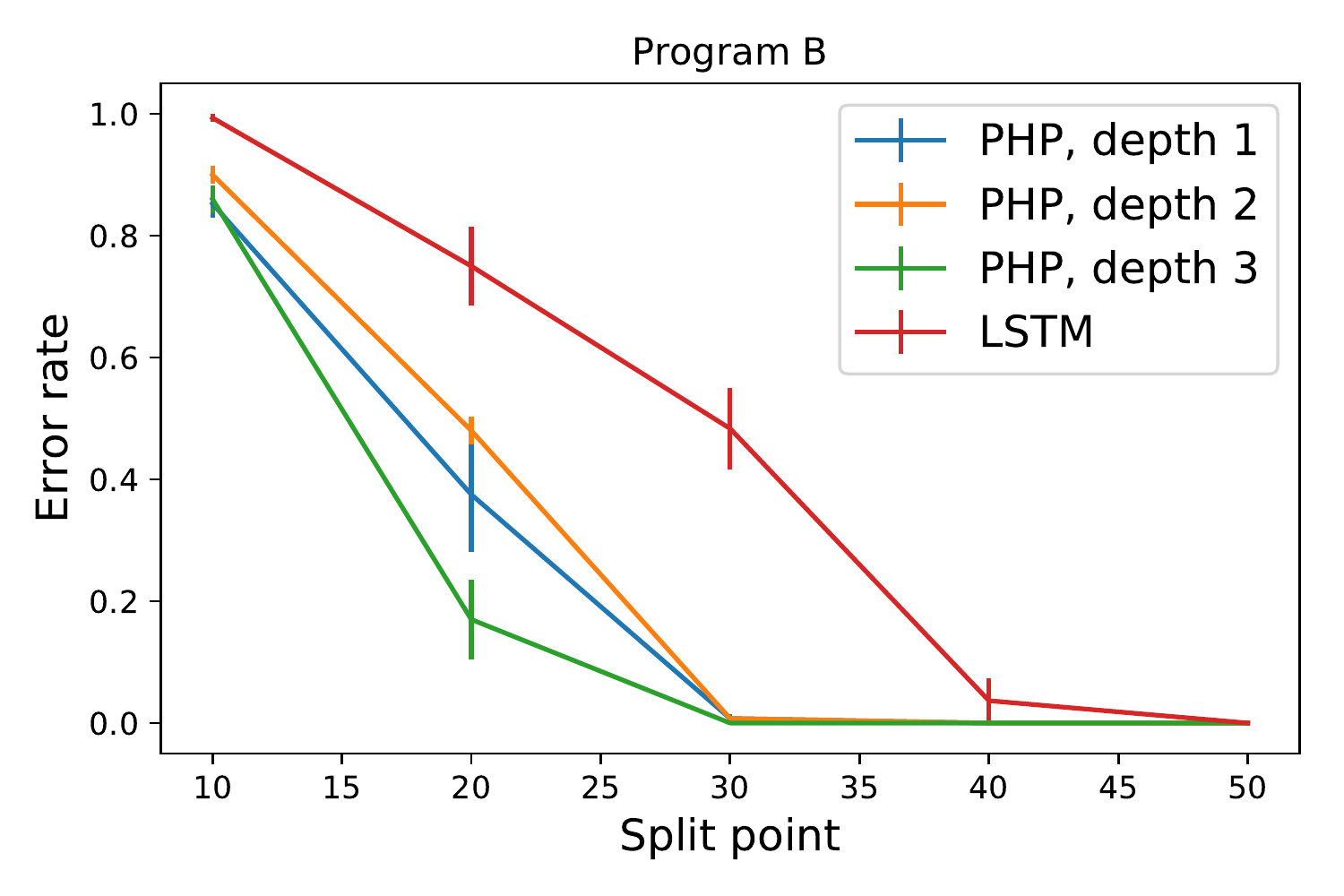}
\hfill
\includegraphics[width=0.29\textwidth]{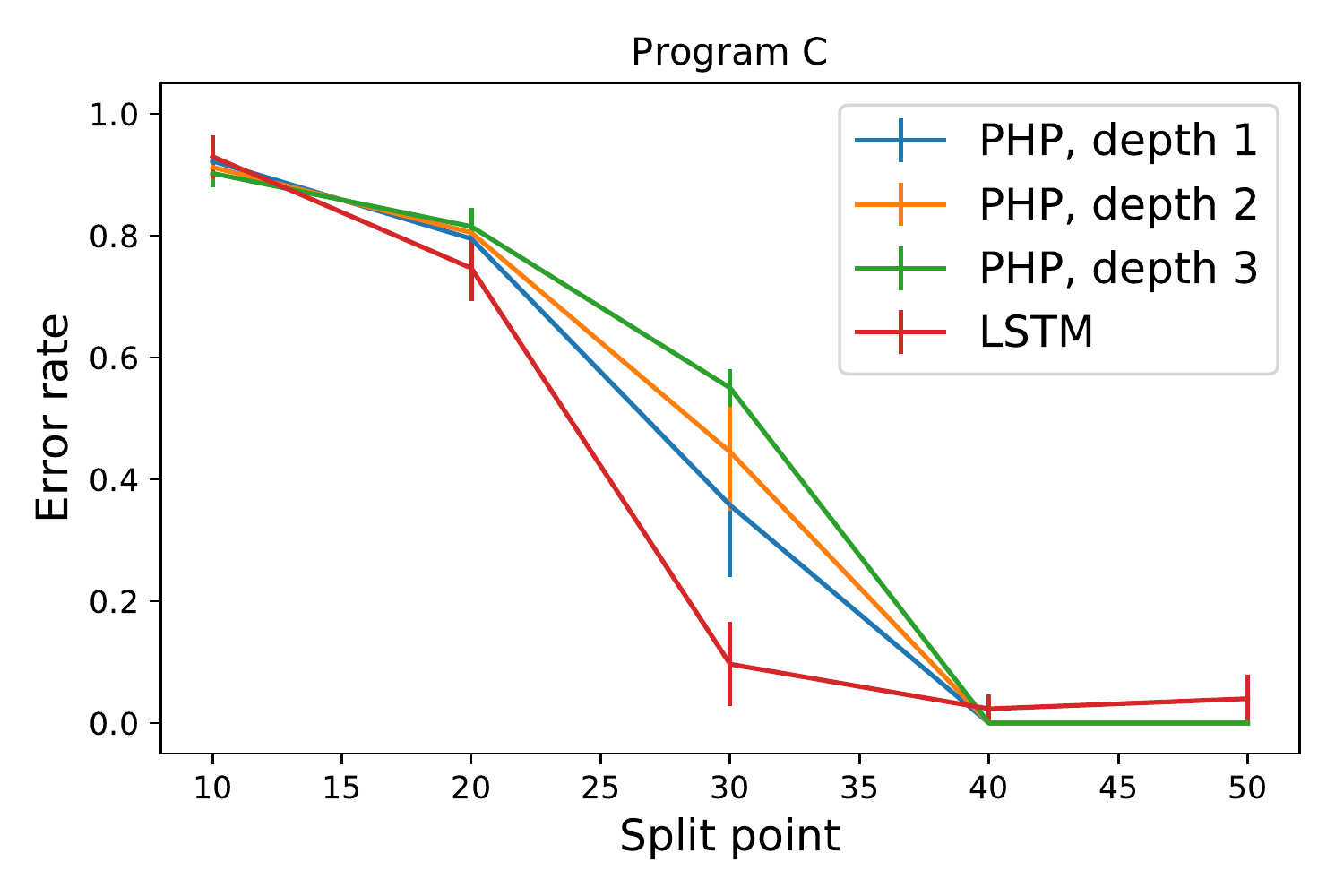}
\hfill
\caption{Error rate in imitation learning of Karel programs trained on demonstrations that are shorter than the test data. As the split point increases from 10\% to 50\%, longer demonstrations are included in the training dataset; however, test demonstrations are always longer still. Except on program C, the hierarchical control policies exhibit better generalization. Results for more Karel programs are included in~\xref{results}.}
\flab{karel-generalization}
\end{figure*}

In these experiments, we compare the performance of the four models on the ability to generalize to inputs requiring longer execution times than those seen during training.
From the 10,000 demonstrations for each program, we computed the set of unique demonstration lengths seen throughout those demonstrations.
For each set of demonstration lengths, we computed the 10th, 20th, 30th, 40th, and 50th percentile lengths.
For each Karel program and for each percentile, we create a training and test set where the training set consists of 250 demonstrations with length less than that percentile trace length, and the test set consists of 100 demonstrations with length at least that percentile trace length.
Thus, each test demonstration is longer than every training demonstration.
\fref{karel-generalization} summarizes the results.

On all but program C, our results suggest that the LSTM exhibits poorer generalization to longer test demonstrations than HVIL.
We hypothesize that the hierarchical structure of PHPs acts as a prior which makes it more likely that they will learn the correct procedures of the program, which can perform reliably on longer demonstrations than ever seen in training.

\section{Conclusion}

In this paper we proposed HVIL, a variational inference method for training hierarchical control policies from demonstrations in which the structure is latent, including a novel architecture for the inference model.
We detailed variance reduction methods that we found necessary for successful training.
We also identified a novel benefit of HVIL, namely the ability to leverage acausal information in decomposing a task into a hierarchy of simpler procedures.

Our method can benefit from further innovation both in variational inference and more generally in stochastic optimization.
We expect that introducing control variates, such as RELAX~\cite{grathwohl2017backpropagation}, can significantly reduce the variance of the gradient estimators and improve convergence rates and robustness.
Tighter bounds on the log-likelihood objective, such as IWAE~\cite{burda2015importance}, can also help in cases where $q_\phi$ is insufficiently expressive to match the posterior, however an open question is how to combine such bounds with the variance reduction methods of~\sref{tech}.

Training hierarchical policies with HVIL opens up the possibility of extending the model with components that enlarge the latent space but provide more expressiveness and regularization.
For example, the procedures of a PHP can take arguments and terminate with return values~\cite{Fox2019Multi}.
PHP would also be more expressive if they were augmented with a recurrent state.

In this work, we refrained from supplying meaningful prior knowledge in the form of a call-graph of procedures.
In future work, we will explore the benefits of a compatible call-graph in improving data efficiency and generalization, and study data-driven methods for assisting the discovery of such call-graphs.

\section*{Acknowledgements}

This research was performed at the RISELab at UC Berkeley in affiliation with the AUTOLAB and Berkeley AI Research (BAIR).
This research was supported in part by: the NSF CISE Expeditions Award CCF-1730628; the NSF National Robotics Initiative Award 1734633; and the Scalable Collaborative Human-Robot Learning (SCHooL) Project.
The authors were supported in part by donations from Alibaba, Amazon Web Services, Ant Financial, Arm, Autodesk, CapitalOne, Comcast, Ericsson, Facebook, Google, Hewlett-Packard, Honda, Huawei, Intel, Knapp, Microsoft, Nvidia, Scotiabank, Siemens, Splunk, Toyota Research Institute, and VMware.

\bibliography{hvi}
\bibliographystyle{icml2019}

\clearpage
\appendix
\section{List of Karel Programs}
\xlab{programs}

\begin{enumerate}
\item Program A: \texttt{def run() \{ turnRight(); while (noMarkersPresent()) \{ move(); if (rightIsClear()) \{ turnRight(); \} \} \} }
\item Program B: \texttt{def run() \{ turnRight(); while (noMarkersPresent()) \{ move(); if (leftIsClear()) \{ turnLeft(); \} \} \} }
\item Program C: \texttt{def run() \{ turnRight(); while (noMarkersPresent()) \{ if (rightIsClear()) \{ turnRight(); \} move(); \} \} }
\item Program D: \texttt{def run() \{ turnRight(); while (noMarkersPresent()) \{ if (frontIsClear()) \{ move(); \} else \{ turnLeft(); \} \} \} }
\item Program E: \texttt{def run() \{ turnRight(); while (noMarkersPresent()) \{ if (frontIsClear()) \{ move(); \} else \{ turnRight(); \} \} \} }
\item Program F: \texttt{def run() \{ turnRight(); while (noMarkersPresent()) \{ if (frontIsClear()) \{ move(); \} else \{ if (rightIsClear()) \{ turnRight(); \} else \{ turnLeft(); \} \} \} \} }
\end{enumerate}

\clearpage
\section{All Karel Experiment Results}
\xlab{results}

\begin{figure*}[b]
\centering
\includegraphics[width=0.32\textwidth]{figures/karel/samplesize_hoc14a.pdf}
\includegraphics[width=0.32\textwidth]{figures/karel/samplesize_hoc15a.pdf}
\includegraphics[width=0.32\textwidth]{figures/karel/samplesize_hoc16a.pdf}
\includegraphics[width=0.32\textwidth]{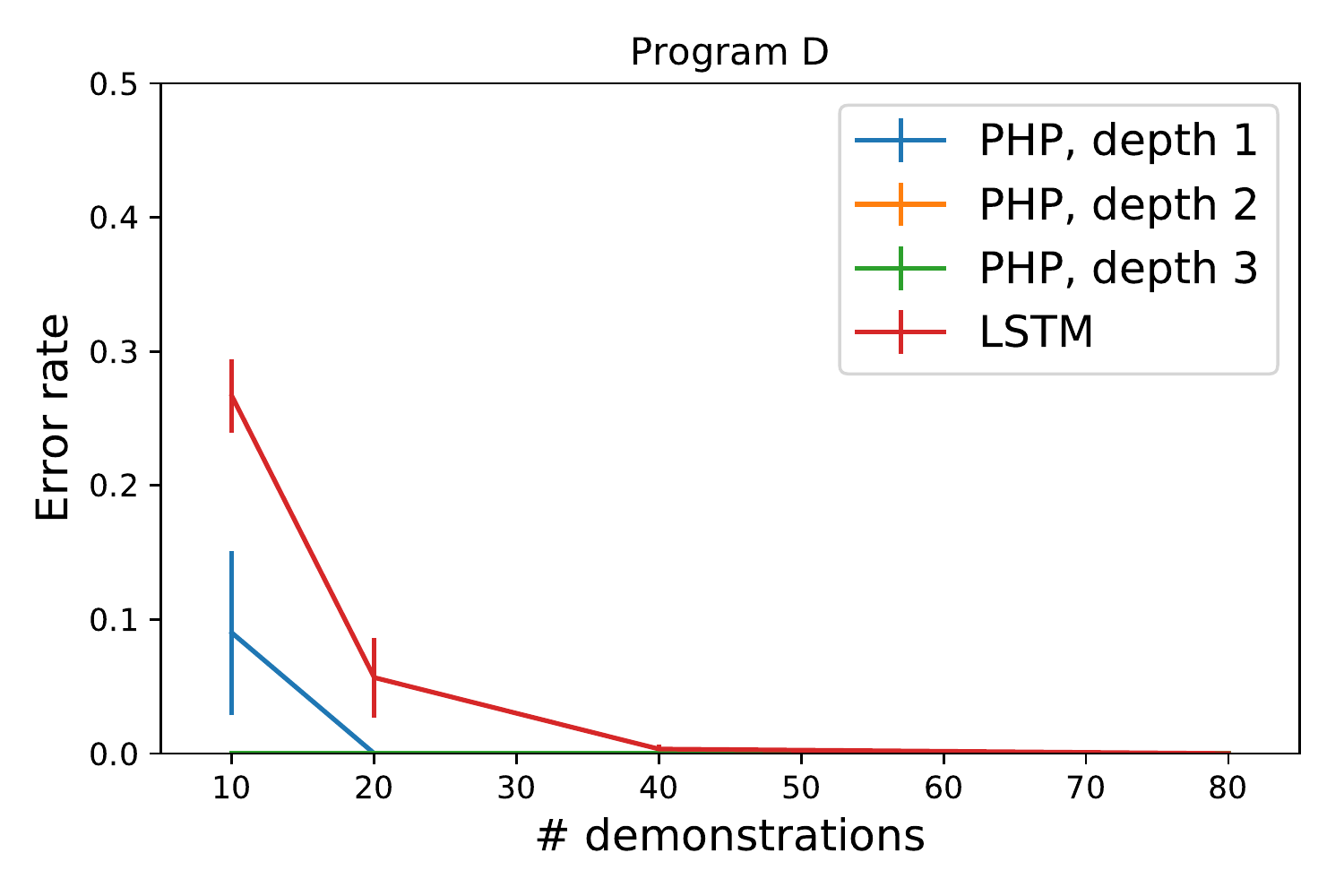}
\includegraphics[width=0.32\textwidth]{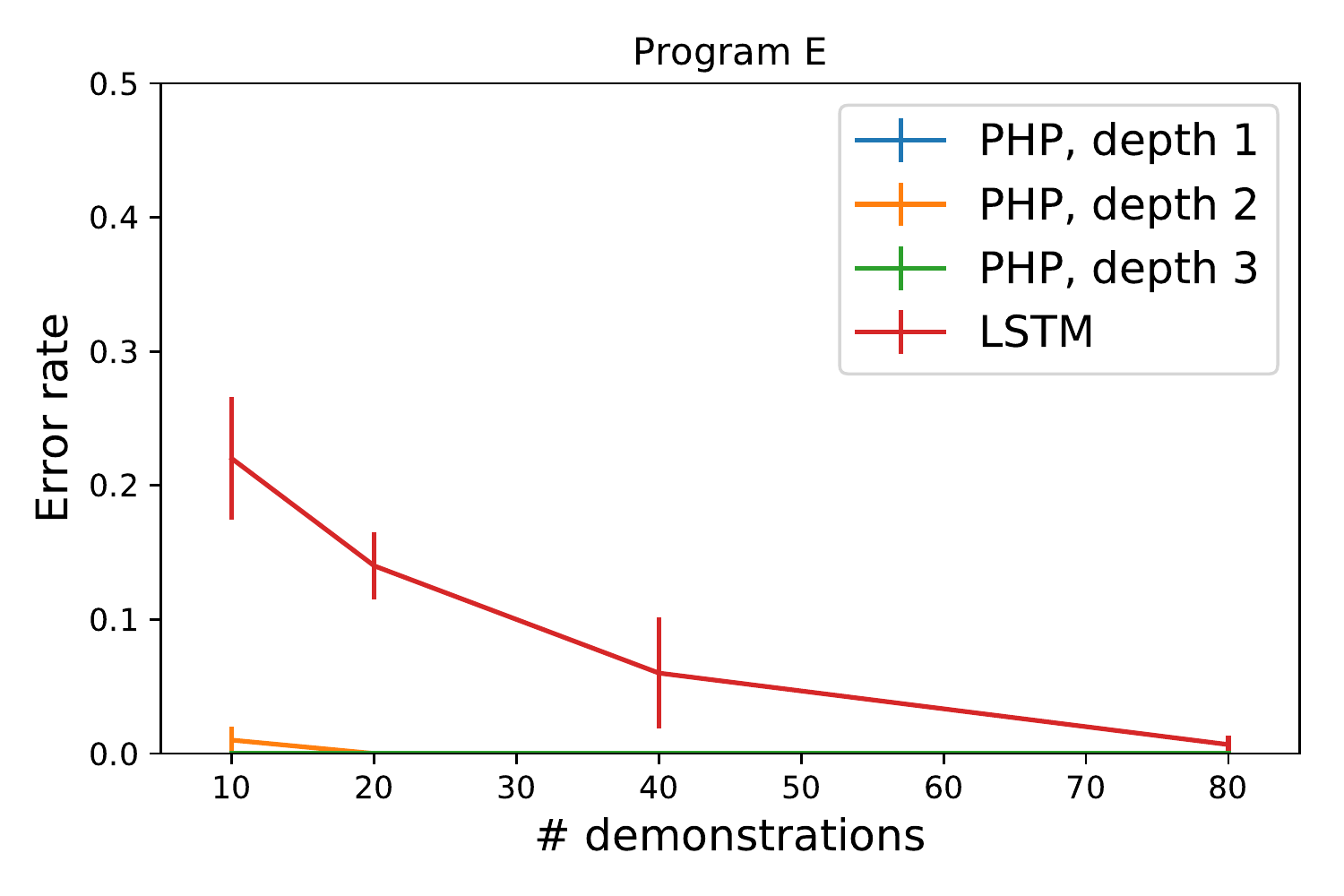}
\includegraphics[width=0.32\textwidth]{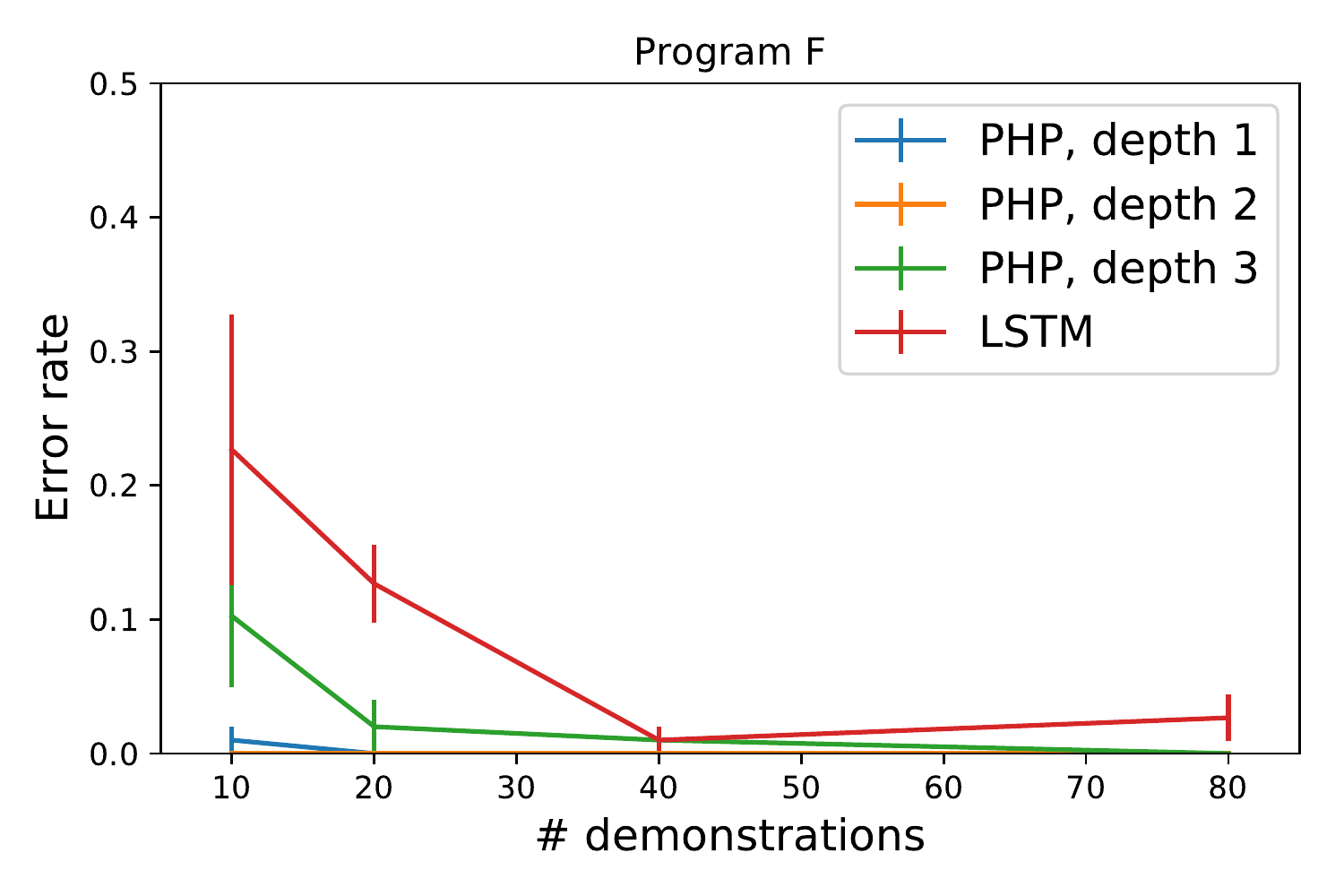}
\caption{Error rate in imitation learning of Karel programs for datasets of training demonstrations of sizes 10 to 80.}
\vspace{3em}
\end{figure*}

\begin{figure*}[b]
\centering
\includegraphics[width=0.32\textwidth]{figures/karel/generalization_hoc14a.pdf}
\includegraphics[width=0.32\textwidth]{figures/karel/generalization_hoc15a.pdf}
\includegraphics[width=0.32\textwidth]{figures/karel/generalization_hoc16a.pdf}
\includegraphics[width=0.32\textwidth]{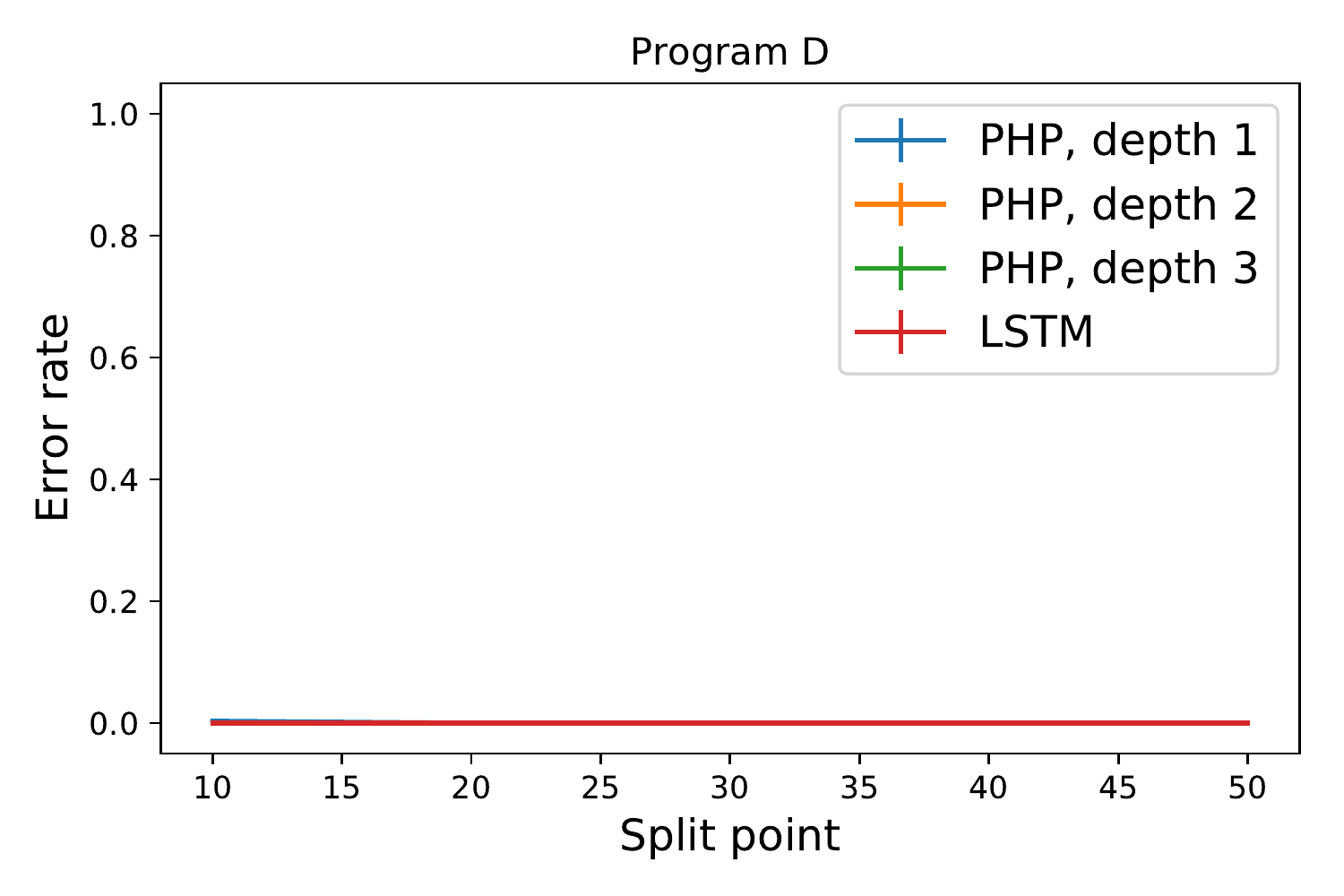}
\includegraphics[width=0.32\textwidth]{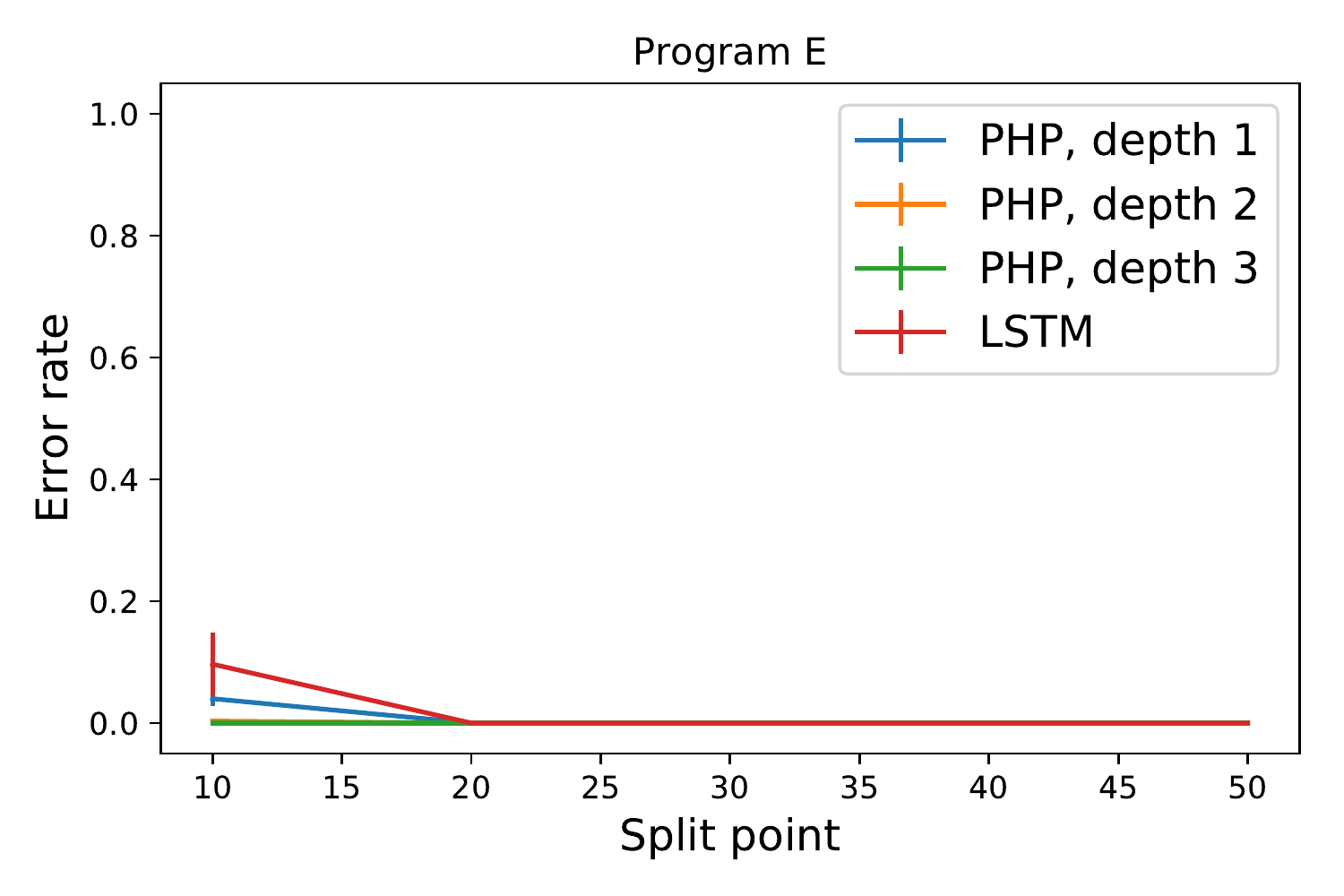}
\includegraphics[width=0.32\textwidth]{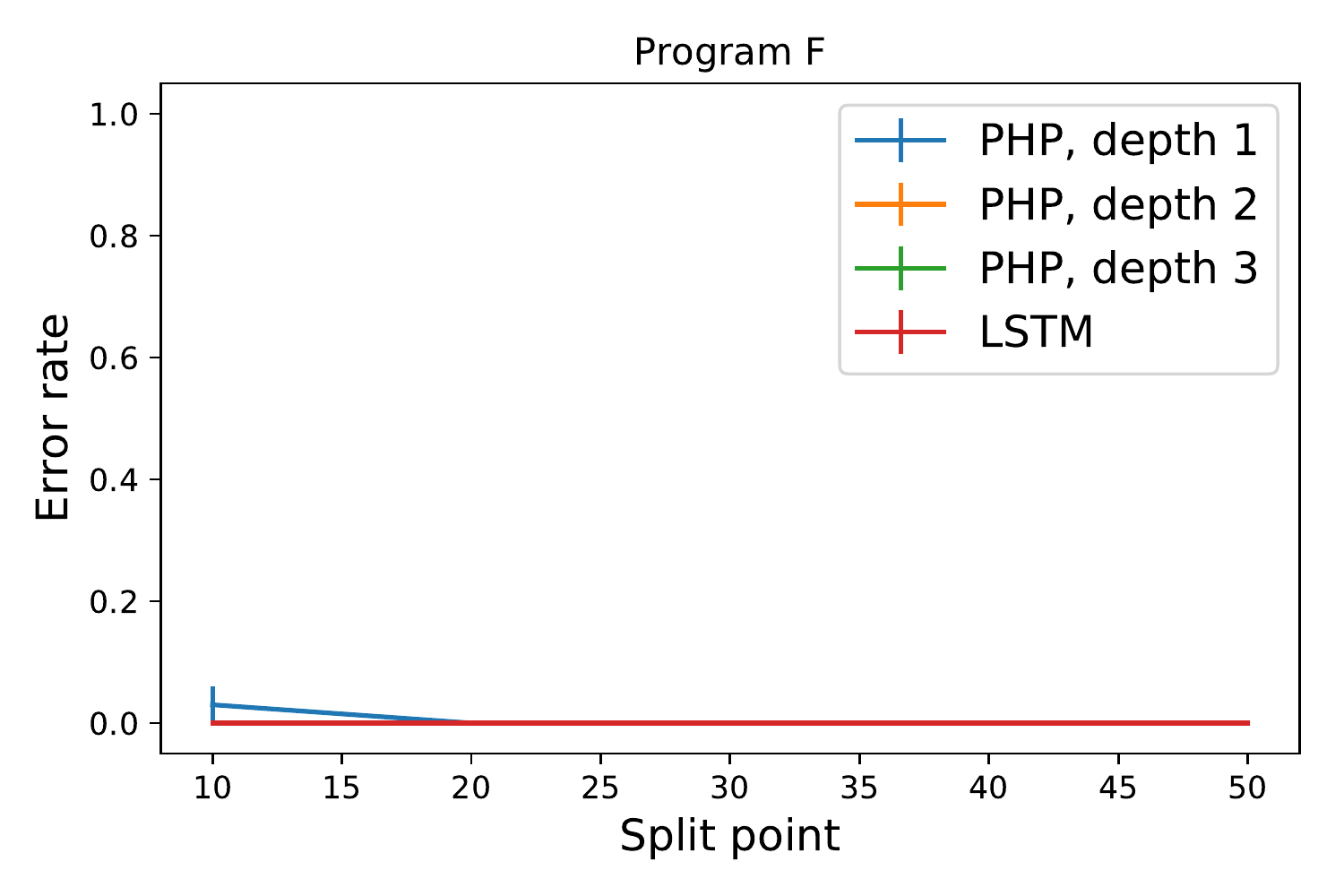}
\caption{Error rate in imitation learning of Karel programs trained on demonstrations that are shorter than the test data.}
\vspace{3em}
\end{figure*}

\end{document}